\documentclass[10pt,twocolumn,letterpaper]{article}

 \usepackage{cvpr}              

\definecolor{cvprblue}{rgb}{0.21,0.49,0.74}
\usepackage[pagebackref,breaklinks,colorlinks,allcolors=cvprblue]{hyperref}
\usepackage{arydshln}
\usepackage{color, colortbl}
\usepackage{xcolor}
\definecolor{gray}{HTML}{E6E6E6}
\definecolor{blue}{HTML}{e8f0f8}
\usepackage{ulem}
\usepackage{amssymb}
\usepackage{array}
\usepackage{xcolor}
\usepackage{booktabs}
\usepackage{multirow}
\usepackage{xcolor}
\usepackage{graphicx}
\usepackage{subcaption} 

\title{Mamba as a Bridge: Where Vision Foundation Models Meet Vision Language Models for Domain-Generalized Semantic Segmentation}

\author{
	Xin Zhang$^{1}$ \quad Robby T. Tan$^{1,2}$ \\
	$^1$National University of Singapore \quad $^2$ASUS Intelligent Cloud Services \\
	{\tt\small x.zhang@u.nus.edu \quad robby.tan@nus.edu.sg}
}

\begin{document}
\maketitle

\begin{abstract}
	
	Vision Foundation Models (VFMs) and Vision-Language Models (VLMs) have gained traction in Domain Generalized Semantic Segmentation (DGSS) due to their strong generalization capabilities~\footnote{In this paper, we refer to foundation models trained solely on visual data as VFMs and those trained on both visual and textual data as VLMs.}. 
	However, existing DGSS methods often rely exclusively on either VFMs or VLMs, overlooking their complementary strengths. 	
	VFMs (e.g., DINOv2) excel at capturing fine-grained features, while VLMs (e.g., CLIP) provide robust text alignment but struggle with coarse granularity. 	
	Despite their complementary strengths, effectively integrating VFMs and VLMs with attention mechanisms is challenging, as the increased patch tokens complicate long-sequence modeling.
	To address this, we propose MFuser, a novel Mamba-based fusion framework that efficiently combines the strengths of VFMs and VLMs while maintaining linear scalability in sequence length.
	MFuser consists of two key components: MVFuser, which acts as a co-adapter to jointly fine-tune the two models by capturing both sequential and spatial dynamics; and MTEnhancer, a hybrid attention-Mamba module that refines text embeddings by incorporating image priors.
	Our approach achieves precise feature locality and strong text alignment without incurring significant computational overhead. 
	Extensive experiments demonstrate that MFuser significantly outperforms state-of-the-art DGSS methods, achieving 68.20 mIoU on synthetic-to-real and 71.87 mIoU on real-to-real benchmarks. The code is available at \url{https://github.com/devinxzhang/MFuser}.
\end{abstract}

\section{Introduction}
\label{sec:intro}

\begin{figure}[th]
	\centering
	\includegraphics[width=0.95\linewidth]{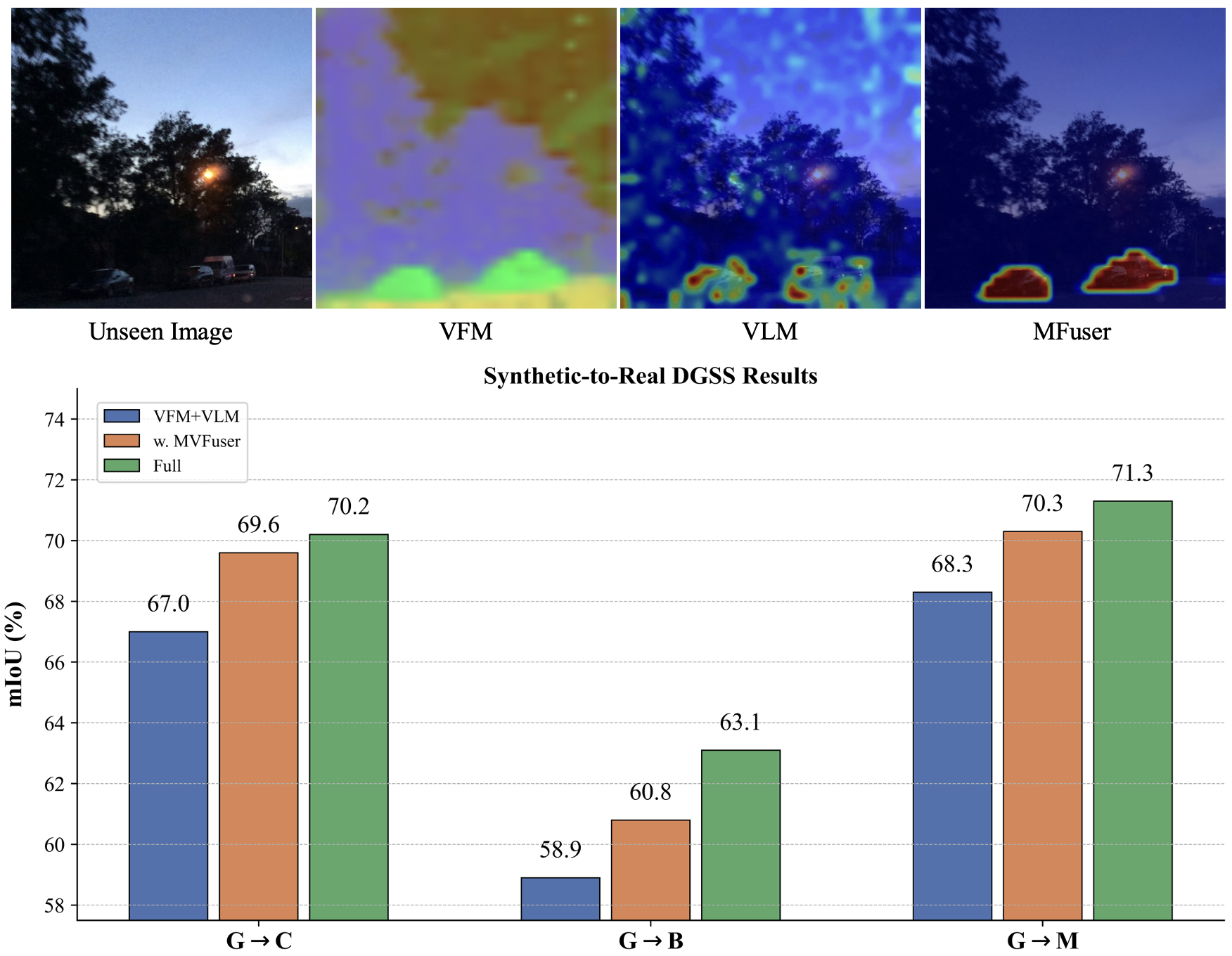}
	\caption{Comparative analysis of the VFM and the VLM features. VFM: Visualization of PCA-computed features from DINOv2 (the first three components of PCA, computed on the image features, serve as color channels), displaying fine-grained details but lacking text alignment. VLM: Image-text similarity map from EVA02-CLIP using the query `car', demonstrating good alignment with text but insufficient localization of queried objects. MFuser: Our proposed fusion framework integrates VFM and VLM, resulting in unified features that exhibit both precise locality and robust text alignment. Quantitative results on synthetic-to-real DGSS benchmarks further validate our approach, with MFuser consistently achieving the highest mIoU scores across all tasks.}
	\label{fig:vfm_vs_vlm}
\end{figure}

Developing semantic segmentation models that can robustly handle diverse and unseen conditions~\cite{chen2023class,yang2025end,yang2024semantic,yang2022object} is critical for real-world applications such as autonomous driving, where variations in environment, lighting, and weather~\cite{lin2024nightrain,lin2024nighthaze,chen2024dual,Yan_2023_ICCV,ai2024domain} can significantly impact performance. Domain Generalized Semantic Segmentation (DGSS) aims for strong performance across unseen domains without relying on target domain data during training. Traditional approaches include normalization and whitening techniques~\cite{choi2021robustnet,peng2022semantic}, domain randomization methods~\cite{huang2023style,zhao2022style,zhong2022adversarial}. Despite these efforts, existing approaches remain suboptimal, as they often rely on conventional backbones pre-trained on limited datasets, which struggle to generalize effectively to the diverse challenges encountered in real-world scenarios.

The recent emergence of Vision Foundation Models (VFMs) and Vision Language Models (VLMs) has established them as powerful tools for achieving generalization in various domains. Some studies have introduced parameter-efficient fine-tuning (PEFT) methods that effectively adapt these foundation models for DGSS~\cite{wei2024stronger,yi2024learning}. Additionally, some works leverage diffusion models~\cite{rombach2022high} to generate diverse-style images for training DGSS models~\cite{fahes2024simple}. VLMs, in particular, have demonstrated the ability to generalize effectively across varied domains by utilizing text embeddings that provide semantic and domain-invariant representations~\cite{radford2021learning}. This capability has sparked the development of multiple approaches in both image classification~\cite{cho2023promptstyler,huang2023sentence} and semantic segmentation~\cite{fahes2024simple,pak2024textual}. However, the specific differences between VFMs and VLMs in the context of DGSS remain underexplored.

VFM features (e.g., DINOv2~\cite{oquab2023dinov2}) capture strong details at a granular level. In contrast, VLM features (e.g., EVA02-CLIP~\cite{fang2024eva}) struggle to associate text semantics with precise visual regions due to their image-level alignment training. However, this alignment enables VLMs to leverage text embeddings as semantic anchors~\cite{pak2025textual}, guiding visual features to remain robust across domain variations. To examine their properties, we perform principal component analysis (PCA) on the DINOv2 features at the final layer. As illustrated in Fig.~\ref{fig:vfm_vs_vlm}, the PCA-computed features from DINOv2 clearly distinguish between different objects (e.g., cars and trees), even in low-light conditions. Additionally, we apply EVA02-CLIP with the text query `car'. The activation map also indicates the presence of cars but appears incomplete. This raises an important question: \textsl{how can we combine both models to extract features that are both locally precise and text-aligned, enabling effective use of text embeddings for improved generalization?}

An intuitive idea would be to utilize both a VFM and a VLM for training a segmentation model. However, without fine-tuning, foundation models may struggle to adapt to DGSS tasks~\cite{wei2024stronger} and VLM text embeddings often fail to align with VFM features, resulting in suboptimal performance. Fully fine-tuning both models, meanwhile, is computationally prohibitive. As such, we propose to introduce additional trainable parameters while keeping the original ones frozen, enabling efficient adaptation. Moreover, combining features from both encoders doubles the patch sequence length, complicating even parameter-efficient fine-tuning methods in handling such long-range sequences. This leads us to our second question: \textsl{how can we efficiently adapt and integrate both a VFM and a VLM for DGSS?}

To this end, we propose MFuser, a novel fusion framework based on the State-Space Model (SSM) that efficiently unifies the strengths of VFMs and VLMs. SSMs~\cite{gu2023mamba,zhu2024vision} are well-suited for capturing long-range dependencies with linear computational complexity, making them ideal for jointly adapting VFMs and VLMs with minimal overhead. Following recent advances in text-guided segmentation~\cite{zhou2023zegclip,ye2024otseg,pak2024textual}, we build MFuser on the text-queried Mask2Former~\cite{cheng2022masked} pipeline, where class text embeddings serve as queries for the segmentation decoder, enabling class-aware feature refinement. Specifically, we introduce \textbf{MV}Fuser, a \textbf{M}amba-based co-adapter that jointly fine-tunes the two \textbf{V}isual models. By taking concatenated patch tokens (features) from both models at each layer, MVFuser models both sequential dynamics and spatial relationships among tokens in parallel. This enables effective interaction between the two feature types, enhancing the granularity of VLM features while also reducing trainable parameters.

To further ensure cross-modality consistency between the fused visual features and VLM \textbf{T}ext embeddings, we introduce \textbf{MT}Enhancer. MTEnhancer employs a hybrid attention-\textbf{M}amba architecture, leveraging the strengths of both model families. Visual features are used as conditional inputs within MTEnhancer, enabling effective sequence modeling that produces text embeddings closely related to visual content, resulting in image-conditioned text embeddings. Extensive experiments across diverse DGSS settings demonstrate that the proposed MFuser consistently outperforms existing state-of-the-art methods, achieving superior results in both synthetic-to-real and real-to-real scenarios. Contributions can be summarized into three aspects:
\begin{itemize}
	\item We propose a novel fusion framework, MFuser, to collaborate arbitrary pairs of VFMs and VLMs for DGSS, integrating the strengths of both without introducing significant computational overhead. 
	\item We present MVFuser, a Mamba-based co-adapter that enables joint fine-tuning of VFMs and VLMs, bridging the gap between these models and enhancing their complementary feature interactions. Additionally, we introduce MTEnhancer, a hybrid attention-Mamba module that refines text embeddings with visual priors, ensuring superior cross-modal consistency and robust alignment.	
	\item Extensive experiments show the proposed MFuser consistently outperforms state-of-the-art methods, achieving 68.20 mIoU on synthetic-to-real and 71.87 mIoU on real-to-real benchmarks.
\end{itemize}

\section{Related Works}
\label{sec:formatting}

\paragraph{Domain Generalized Semantic Segmentation}
Domain Generalized Semantic Segmentation (DGSS) aims to develop models capable of generalizing to unseen domains without relying on target domain data during training. Common approaches include meta-learning, which exposes models to diverse tasks to learn features that are robust to domain shifts~\cite{kim2022pin}; data augmentation techniques, such as style transfer and synthetic data creation, to introduce extensive visual diversity~\cite{chattopadhyay2023pasta};
instance normalization and whitening~\cite{peng2022semantic,xu2022dirl,huang2019iterative,pan2018two}, which encourages the model to foucs on domain-invariant features rather than domain-specific styles. Some works also explore to design new architectures based on transformers~\cite{ding2023hgformer,hoyer2022hrda}. Recently, increasing attention has been paid to leveraging foundation models to enhance generalization~\cite{wei2024stronger,yi2024learning,pak2025textual}. Efforts have been taken to harness generative foundation models to creat new images~\cite{benigmim2024collaborating}, parameter-efficiently fine-tune VFMs~\cite{wei2024stronger}, leverage textual semantics to guide invariance learning~\cite{pak2025textual}, etc. However, the complementary potential of combining VFMs and VLMs remains largely underexplored.

\vspace{-3mm}
\paragraph{Foundation Models}
Foundation models represent a transformative approach in deep learning, focusing on pre-training networks on a vast collection of unlabeled images. This pre-training equips the model with strong general representation capabilities, allowing it to be fine-tuned effectively for various downstream tasks. Initially popularized in Natural Language Processing (NLP), this paradigm has also drawn increasing attention in computer vision. In this paper, we refer to the vision-only pre-trained models as Vision Foundation Models (VFMs) including DINO~\cite{caron2021emerging} and DINOv2~\cite{oquab2023dinov2}, iBOT~\cite{zhou2021ibot}, MAE~\cite{he2022masked}, SAM~\cite{kirillov2023segment}, etc. Vision-language pre-trained models are referred to as Vision Language Models (VLMs), which include CLIP~\cite{radford2021learning}, EVA02-CLIP~\cite{fang2024eva,sun2023eva}, SIGLIP~\cite{zhai2023sigmoid}, etc. There are also generative foundation models such as Stable Diffusion~\cite{rombach2022high,wu2024towards}. We focus on effectively combining VFMs and VLMs for DGSS. 

\vspace{-3mm}
\paragraph{State Space Models for Visual Applications}
State-space models (SSMs)~\cite{gu2021efficiently,smith2022simplified} have emerged as promising alternatives for capturing long-range dependencies, offering linear scalability with sequence length. Building on the foundational S4 model~\cite{gu2021efficiently}, which introduced deep state-space modeling, SSMs have found applications across a range of fields, including Natural Language Processing (NLP)~\cite{mehta2022long}, computer vision~\cite{zhu2024vision}, medical applications~\cite{ruan2024vm}. Mamba~\cite{gu2023mamba} extended S4 by introducing a hardware-aware design and a selective scan mechanism, leading to the development of a selective SSM called the S6 model. More recently, VMamba~\cite{zhu2024vision} emerged as a fully Mamba-based architecture for vision tasks, while other studies~\cite{hatamizadeh2024mambavision} explored hybrid models combining Mamba and transformers. Unlike previous SSM-based efforts that primarily focus on creating entire backbone architectures, we take a different approach by designing Mamba-based adapters to efficiently fine-tune pre-trained VFMs and VLMs. This method enhances the adaptability and performance of VFMs and VLMs across various domains, leveraging Mamba’s efficiency to optimize existing models rather than training from scratch.

\section{Preliminary}
\vspace{-1mm}
\paragraph{Domain Generalized Segmantic Segmentation}
Given the source images $\mathcal{X}^{S}=\{x_{i}^{S}\}_{i=1}^{N_{S}}$ with corresponding ground truth masks $\mathcal{Y}^{S}=\{y^{S}_{i}\}^{N_{S}}_{i=1}$ where $N_{S}$ denotes the number of source images, and a segmentation model $M$, composed of a visual encoder $E$ followed by a segmentation decoder $D$, namely $M=D\circ E$, domain generalized semantic segmentation (DGSS) aims to train the network to generalize to unknown target domains. With the advancements in foundation models, recent DGSS methods increasingly leverage their strong generalization capabilities to design effective visual encoders~\cite{wei2024stronger,yi2024learning}.

\vspace{-3mm}
\paragraph{Semantic Segmentation with Text Queries}
Recent segmentation frameworks like Mask2Former~\cite{cheng2022masked}, utilize a query-based mechanism where learnable object queries serve as dynamic pointers to direct the model's focus on relevant regions. Building on this, recent studies have increasingly leveraged the image-text alignment capabilities of Vision Language Models (VLMs) to design text-based queries~\cite{zhou2023zegclip,ye2024otseg,pak2024textual,das2024mta,liu2023grounding,li2023semantic,lei2024ez}. The text embeddings produced by VLMs have been found to be inherently domain-invariant, capturing semantic information that remains consistent across various contexts and visual styles. This domain invariance stems from the VLM training process, which associates textual descriptions with diverse visual inputs, effectively disentangling semantic content from domain-specific features. The domain invariance of text embeddings forms a basis for promoting the domain generalization of visual features. In this paper, we follow a similar pipeline which utilizes the text embeddings of each class as the queries in a Mask2Former decoder. Formally, the visual encoder $E^{\rm VLM}_{V}$ of a VLM serves as the encoder of the segmentation model, the aligned text encoder $E^{\rm VLM}_{T}$ generates class embeddings $q_t=[t^{1}, t^{2}, ..., t^{C}]$ for each class label name $\{\rm class_k\}^{C}_{k=1}$. $q_t$ will be used to design queries or conditional queries of the decoder~\cite{zhou2023zegclip,ye2024otseg,pak2024textual}. 

\begin{figure*}[t]
	\centering
	\includegraphics[width=0.96\linewidth]{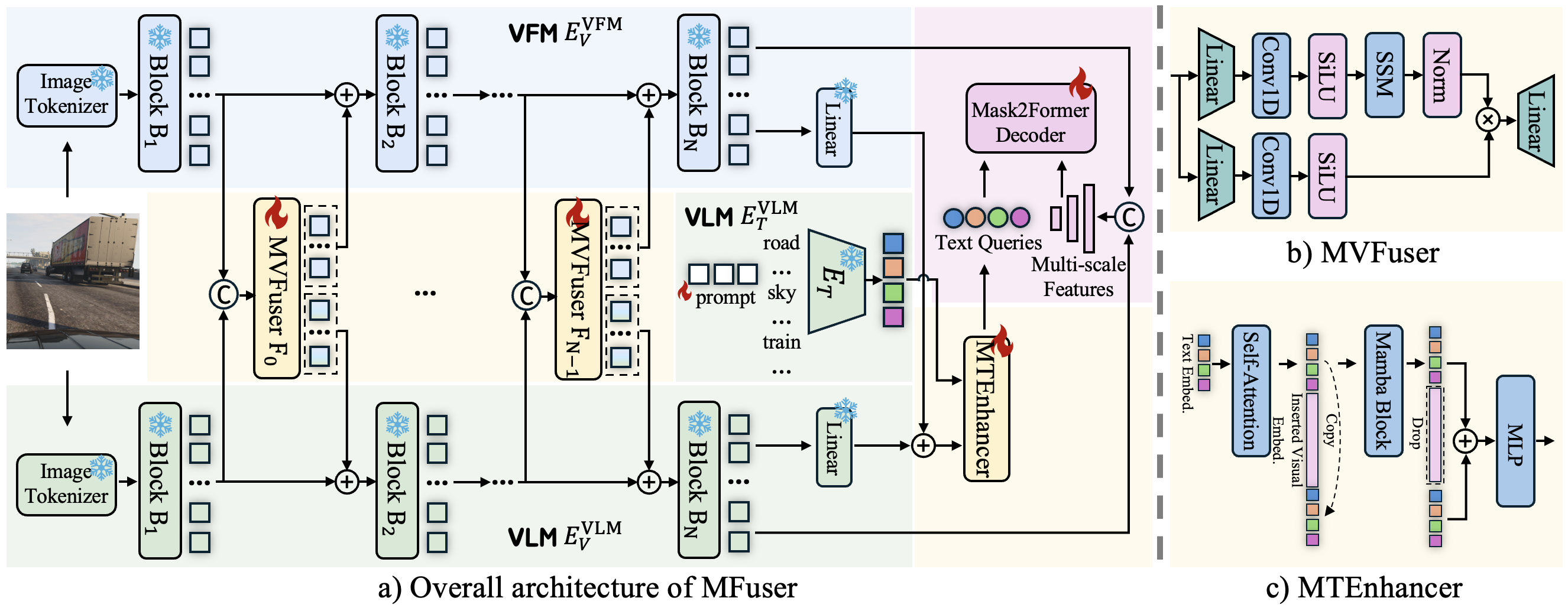}
	\caption{Overall architecture of MFuser. MFuser takes inputs through both VFM and VLM visual encoders. Features from each encoder layer are concatenated and refined in MVFuser, which captures sequential and spatial dependencies in parallel. The refined features are then added back to the original features and passed to the next layer. MTEnhancer strengthens text embeddings of each class by integrating visual features through a hybrid attention-Mamba mechanism. The enhanced text embeddings serve as object queries for the Mask2Former decoder, alongside multi-scale visual features. During training, only MVFusers, MTEnhancers, and the segmentation decoder are trainable while the VFM and VLM remain frozen, preserving their generalization ability and enabling efficient training. Note that skip connections between each block of MTEnhancer are omitted for clarity.}
	\label{fig:main}
\end{figure*}

\section{Proposed Method}
\vspace{-1mm}
In this section, we introduce the Mamba-based foundation models fuser (MFuser), a framework designed to integrate an arbitrary VFM with a CLIP-like VLM using a Mask2Former decoder for DGSS. Fig.~\ref{fig:main} illustrates the overall architecture of MFuser. MFuser enhances feature locality while leveraging domain-invariant semantic knowledge provided by text embeddings to effectively constrain visual representations. The core components of this framework include MVFuser and MTEnhancer. MVFuser jointly fine-tunes the visual encoders of both models in a parameter-efficient manner, fusing their features to maximize synergy. MTEnhancer enriches the text queries by incorporating visual features, enhancing semantic alignment and feature robustness. 

\subsection{MVFuser}

Due to the large number of parameters in the VFM and VLM visual encoders, fully fine-tuning all parameters is impractical. Instead, we propose the introduction of additional modules, MVFuser, to refine visual features while keeping the original encoder parameters frozen. 

This design offers several advantages. First, the distinct characteristics of the two visual encoders could be compromised by full fine-tuning, whereas adapter-style fine-tuning preserves their original strengths while mitigating their weaknesses. Second, refining features from both encoders through a shared MVFuser encourages effective interaction between the two feature types.

Specifically, the visual encoders of VFMs and VLMs are composed of an image tokenizer layer and $N$ consecutively connected transformer blocks $\{B_i\}^{N}_{i=1}$. The image tokenizer layer first converts a 2D image into flatten patch tokens $x_p\in\mathbb{R}^{T\times D}$, where $T$ represents the length of the patch sequence and $D$ denotes the feature dimension. 

Normally, $x_p$ is input into the transformer blocks to calculate features. The process is as follows:
\vspace{-2mm}
\begin{equation}
	x_1 = B_1(x_p), x_i = B_i(x_{i-1}),
\end{equation}
where $x_i$ is the token features output by Block $B_i$. The features for VFM and VLM can be denoted as $x_i^{\text{\scriptsize VFM}}$, and $x_i^{\text{\scriptsize VLM}}$, respectively. 

As stated, $x_i^{\text{\scriptsize VFM}}$ exhibits finer granularity, from which $x_i^{\text{\scriptsize VLM}}$ can benefit through the interaction. We propose inserting the MVFuser at each block to bridge the two visual encoders, encouraging layer-wise interaction of the two models. MVFuser receives both $x_i^{\text{\scriptsize VFM}}$ and $x_i^{\text{\scriptsize VLM}}$ as input, the learned feature offsets are then added back to $x_i^{\text{\scriptsize VFM}}$ and $x_i^{\text{\scriptsize VLM}}$, respectively, enabling multi-level feature refinement where one MVFuser refines the features from both encoders:
\begin{align}
	[\Delta x_i^{\text{\scriptsize VFM}}; \Delta x_i^{\text{\scriptsize VLM}}] = \mathrm{MVFuser}([x_i^{\text{\scriptsize VFM}}; x_i^{\text{\scriptsize VLM}}]), \\
	{x_i^{\text{\scriptsize VFM}}}^{\prime} = x_i^{\text{\scriptsize VFM}} + \Delta x_i^{\text{\scriptsize VFM}}, 
	{x_i^{\text{\scriptsize VLM}}}^{\prime} = x_i^{\text{\scriptsize VLM}} + \Delta x_i^{\text{\scriptsize VLM}},
\end{align}
where $\Delta x_i^{\text{\scriptsize VFM}}$ and $\Delta x_i^{\text{\scriptsize VLM}}$ are the learned feature offsets for VFM and VLM, respectively. ${x_i^{\text{\scriptsize VFM}}}^{\prime}$ and ${x_i^{\text{\scriptsize VLM}}}^{\prime}$ symbolize the refined features.

MVFuser acts two roles: 1) refines $x_i^{\text{\scriptsize VFM}}$ and $x_i^{\text{\scriptsize VLM}}$ to generate more task-specific features; 2) interacts between two kinds of features to complement each's weaknesses. A natural idea to capture inter-token relationship is to employ self-attention mechanism. However, the sequence length is doubled with the features from the two encoders. Applying the attention mechanism in transformers for adaptation is inefficient due to the quadratic increase in computational complexity with token count. While introducing learnable tokens and applying cross-attention between learnable tokens and patch token features can reduce this computational cost, it struggles to capture inter-token dependencies effectively. To address these challenges, we design a fusion module based on state-space models for efficient long-range sequence modeling. 

\paragraph{Core of the MVFuser}
The architecture of MVFuser is shown in Fig.~\ref{fig:main}. Token features from both encoders are concatenated to form the input to MVFuser. Following a bottleneck design, MVFuser first projects the concatenated token features to a lower-dimensional space, models inter-token dependencies, and then projects them back to the original feature dimension.

We modify the original Mamba block to encourage the two branches to capture the sequential dynamics and spatial relationships respectively in parallel. 
\vspace{-1mm}
\begin{align}
	x_i^{(\mathrm{seq})} &= \mathrm{SSM}(\mathrm{conv}(\mathrm{proj}([x_i^{\text{\scriptsize VFM}}; x_i^{\text{\scriptsize VLM}}]))), \\
	x_i^{(\mathrm{spa})} &= \mathrm{conv}(\mathrm{proj}([x_i^{\text{\scriptsize VFM}}; x_i^{\text{\scriptsize VLM}}])).
\end{align}
Note that we omit the activation and normalization layers for clarity. Finally, a gating mechanism is applied between the outputs of the two branch to improve generalization, followed by a projection layer to recover the feature dimension. 
\begin{align}
	[\Delta x_i^{\text{\scriptsize VFM}}; \Delta x_i^{\text{\scriptsize VLM}}]  &= \mathrm{proj}(x_i^{(\mathrm{seq})} \otimes x_i^{(\mathrm{spa})}),
\end{align}
where $\otimes$ denotes the element-wise multiplication.

\subsection{MTEnhancer}

Text embeddings have been utilized as queries in semantic segmentation by framing the task as a matching problem between representative class queries and image patch features, or by serving as the initial object queries for the Mask2Former decoder. This approach leverages the domain-invariant semantic information embedded in text to enhance the model's ability to accurately identify and segment relevant regions within an image~\cite{zhou2023zegclip,ye2024otseg}. Unlike previous methods, which typically assume that visual features and text embeddings are already aligned in a pretrained VLM, our approach enhances the original text embeddings from a VLM by incorporating the fused visual priors through the proposed MTEnhancer. MTEnhancer is designed to enriches text embeddings by modeling their relationships with fused image tokens.

As illustrated in Fig.~\ref{fig:main}, MTEnhancer is a hybrid architecture combining an attention block, a conditional Mamba block, and an MLP, leveraging the strengths of diverse model architectures. The attention block encodes inter-class relationships, while the conditional Mamba block integrates image tokens into the text embeddings. While the Mamba block excels at processing long token sequences, its use in cross-attention mechanisms remains largely unexplored. To efficiently leverage the unidirectional scan order inherent to Mamba, we propose concatenating two copies of text embeddings at both sides of the image token, together they serve as the input of the Mamba block. Each block within MTEnhancer is implemented with residual connections.
\begin{align}
	q_t &= q_t + \mathrm{Attention}(q_t), \\
	[\Delta q_t; \Delta x_v; \Delta q_t^{\rm copy}] &= \mathrm{Mamba}([q_t; x_v; q_t^{\rm copy}]), \\
	q_t &= q_t + \Delta q_t + \Delta q_t^{\rm copy}, \\
	q_t &= q_t + \mathrm{MLP}(q_t),
\end{align}
where $x_v$ represents the fused visual features output by the encoders' final heads. $q_t$ is denoted without distinguishing between updates throughout the process. We adopt the approach of using enhanced text embeddings $q_t$ as object queries for a Mask2Former decoder~\cite{pak2024textual,ye2024otseg}.

\paragraph{Training Objective}
We train the framework with the prediction-level segmentation loss together with the feature-level alignment loss. For the segmentation loss, we follow the standard Mask2Former~\cite{cheng2022masked}:
\begin{equation}
	\mathcal{L_\mathrm{seg}} = \lambda_{\mathrm{bce}}\mathcal{L_\mathrm{bce}} + \lambda_{\mathrm{dice}}\mathcal{L_\mathrm{dice}} + \lambda_{\mathrm{cls}}\mathcal{L_\mathrm{cls}},
\end{equation}
where $\mathcal{L_\mathrm{bce}}$, $\mathcal{L_\mathrm{dice}}$, $\mathcal{L_\mathrm{cls}}$ represent the binary cross-entropy loss and the dice loss for the predicted masks, and the cross-entropy loss for each queried proposal, respectively. 

Additionally, we enforce a pixel-level vision-language alignment using a pixel-text alignment loss to ensure that textual semantics are precisely mapped to corresponding image regions~\cite{rao2022denseclip}. The experiments involve three VLMs: CLIP, EVA02-CLIP, and SIGLIP. We apply SoftMax loss for CLIP and EVA02-CLIP, and Sigmoid loss for SIGLIP, consistent with the loss functions used during each VLM's original training. Therefore, the overall training loss is: 
\vspace{-1mm}
\begin{equation}
	\mathcal{L_\mathrm{total}} = \mathcal{L}_\mathrm{seg} + \mathcal{L}_\mathrm{align}.
\end{equation}

\section{Experiments}

\subsection{Settings}
\noindent \textbf{Datasets} 
We evaluate the performance of MFuser on both synthetic-to-real, clear-to-adverse-weather, and real-to-real scenarios are involved. 
As synthetic datasets, GTAV~\cite{richter2016playing} contains 12,403, 6,382, and 6181 images for training, validation, and testing, respectively, at a resolution of 1914$\times$1052.
As real-world datasets, Cityscapes~\cite{cordts2016cityscapes} comprises 2,975 images for training and 500 images for validation, with a resolution of 2048$\times$1024.
BDD100K~\cite{yu2020bdd100k} includes 7,000 and 1,000 images for training and validation, each at 1280$\times$1024 resolution.
Mapillary~\cite{neuhold2017mapillary} consists of 18,000 training and 2,000 validation images, with varying resolutions across the dataset.
We also include the clear-to-adverse-weather generalization in the supplement.

\noindent \textbf{Network Architecture} 
To make a comprehensive evaluation of the proposed MFuser, we employ the VFM of DINOv2~\cite{oquab2023dinov2}, and VLMs including CLIP~\cite{radford2021learning}, EVA02-CLIP~\cite{sun2023eva}, SIGLIP~\cite{zhai2023sigmoid}. For the segmentation decoder, we follow tqdm~\cite{pak2025textual} which modifies a standard Mask2Former decoder by replacing the randomly initialized object queries with the enhanced class embeddings. Thus, the text object queries are set to 19 to match the number of classes.

\noindent \textbf{Implementation Details} We keep the parameters of the VFM and VLM frozen and only train the MVFuser, MTEnhancer and the segmentation decoder. We use the same training configuration on all VLM alternatives and both generalization setups. We also apply prompt tuning for the text encoder, similar to~\cite{pak2025textual}. All experiments are conducted with the input size of 512$\times$512, a batch size of 2 and learning rate of 1e-4. Following~\cite{wei2024stronger,pak2025textual}, AdamW optimizer is employed with a linear warm-up over $t_{\rm warm}=1.5k$ iterations, followed by a linear decay. Standard augmentations for segmentation tasks are applied, including random scaling, random cropping, random flipping, and color jittering. All experiments are conducted on one 24GB RTX A5000. 

\subsection{Comparison with State-of-The-Art Methods}
We compare our MFuser with existing DGSS methods on two setups: synthetic-to-real (G$\rightarrow$\{C, B, M\}) and real-to-real (C$\rightarrow$\{B, M\}). Three VLMs are involved together with DINOv2, namely CLIP, EVA02-CLIP, and SIGLIP, all of \textsl{Large} types. We mainly compare with recent foundation model-based approaches, including CLOUDS~\cite{benigmim2024collaborating}, VLTseg~\cite{hummer2023vltseg}, Rein~\cite{wei2024stronger}, SET~\cite{yi2024learning}, and tqdm~\cite{pak2025textual}. Several conventional methods are also involved. We provide results on Synthia~\cite{ros2016synthia} and ACDC~\cite{sakaridis2021acdc} in the supplement. 

\vspace{-4mm}
\paragraph{Synthetic-to-Real Generalization}
Tab.~\ref{tab: main_s2r} compares the performance of the proposed MFuser with existing state-of-the-art DGSS methods under the synthetic-to-real setup. For each combination of the VFM and VLMs, we consistently outperform the existing methods on all benchmarks by a large margin. In particular, our MFuser with the EVA02-CLIP model improves the G$\rightarrow$B benchmark by 1.49 mIoU. On average, we achieve 2.15 mIoU better than the state-of-the-art. Our proposed MFuser remains excellent performance using different VFM and VLM combinations, showing the versatility of our framework. To better understand how the proposed MFuser improves the feature generalization, Fig.~\ref{fig:quali} shows the qualitative comparison with the most recent methods, Rein~\cite{wei2024stronger} and tqdm~\cite{pak2025textual}. Our method identifies fine-grained differences more effectively.
\begin{table}[t]
\centering
\caption{
Performance comparison (mIoU in \%) under the synthetic-to-real setting (G$\rightarrow$$\{$C, B, M$\}$). DINOv2~\cite{oquab2023dinov2} is used as the VFM for all MFuser variants, showing only the applied VLMs. Our method is marked in \colorbox{gray}{gray}. The best and second-best results are highlighted in \textbf{bold} and \underline{underlined}, respectively.}
\resizebox{1\columnwidth}{!}{
{\begin{tabular}{l|l|c|c|c|c}
\toprule
\multirow{2}{4em}{Method} & \multirow{2}{3em}{Backbone} & \multicolumn{4}{c}{\textmd{\textbf{synthetic-to-real}}} \\
   &   & G$\rightarrow$C & G$\rightarrow$B & G$\rightarrow$M & Avg. \\ 
\midrule
SAN-SAW~\cite{peng2022semantic} & RN101 & 45.33 & 41.18 & 40.77 & 42.43 \\ 
WildNet~\cite{lee2022wildnet} & RN101 &  45.79 & 41.73 & 47.08 & 44.87 \\
SHADE~\cite{zhao2022style} & RN101 & 46.66 & 43.66 & 45.50 & 45.27 \\ 
TLDR~\cite{kim2023texture} & RN101  & 47.58 &  44.88 & 48.80 & 47.09 \\  
FAMix~\cite{fahes2023simple} & RN101 & 49.47 & 46.40 & 51.97 & 49.28 \\ 
\midrule
SHADE~\cite{zhao2023style} & MiT-B5 & 53.27 & 48.19 & 54.99 & 52.15 \\ 
IBAFormer~\cite{sun2023ibaformer} & MiT-B5 & 56.34 & 49.76 & 58.26 & 54.79 \\ 
VLTSeg~\cite{hummer2023vltseg} & CLIP-B & 47.50 & 45.70 & 54.30 & 49.17 \\ 
\midrule
CLOUDS~\cite{benigmim2024collaborating} & ConvNeXt-L & 60.20 & 57.40 & 67.00 & 61.50 \\
VLTSeg~\cite{hummer2023vltseg} & EVA02-L & 65.60 & 58.40 & 66.50 & 63.50 \\
Rein~\cite{wei2024stronger}  &  EVA02-L & 65.30 & 60.50 & 64.90 & 63.60 \\
Rein~\cite{wei2024stronger} & DINOv2-L & 66.40 & 60.40 & 66.10 & 64.30 \\
SET~\cite{yi2024learning} & DINOv2-L & 68.06 & 61.64 & 67.68 & 65.79 \\
tqdm~\cite{pak2025textual}  & EVA02-L   & 68.88 & 59.18 & 70.10 & 66.05 \\ 
\rowcolor{gray} MFuser & CLIP-L & \textbf{71.24} & 61.08 & 71.14 & 67.82 \\
\rowcolor{gray}
MFuser & SIGLIP-L & \underline{71.10} & \underline{61.19} & \textbf{71.71} & \underline{68.00} \\
\rowcolor{gray}
MFuser & EVA02-L & 70.19 & \textbf{63.13} & \underline{71.28} & \textbf{68.20} \\
\bottomrule
\end{tabular}}
\label{tab: main_s2r}
}
\end{table}

\vspace{-6mm}
\paragraph{Real-to-Real Generalization}
As shown in Tab.~\ref{tab:main_r2r}, we compare the performance of MFuser with existing state-of-the-art DGSS methods under the real-to-real setting. MFuser largely surpasses the existing methods with all three VLMs. Specifically, we improve the C$\rightarrow$B benchmark by 0.74 mIoU, and the C$\rightarrow$M benchmark by 1.7 mIoU. An overall improvement of 1.43 mIoU is achieved. 
\begin{table}[t]
\centering
\caption{
Performance comparison (mIoU  in \%) under the real-to-real setting (C$\rightarrow$$\{$B, M$\}$). DINOv2~\cite{oquab2023dinov2} is used as the VFM for all MFuser variants, showing only the applied VLMs. Our method is marked in \colorbox{gray}{gray}. The best and second-best results are highlighted in \textbf{bold} and \underline{underlined}, respectively.}
\resizebox{0.95\columnwidth}{!}{
{\begin{tabular}{l|l|c|c|c}
\toprule
\multirow{2}{4em}{Method} & \multirow{2}{4em}{Backbone} & \multicolumn{3}{c}{\textmd{\textbf{real-to-real}}} \\
   &   & B & M & Avg. \\ 
\midrule
SAN-SAW~\cite{peng2022semantic} & RN101 & 54.73 & 61.27 & 58.00 \\ 
WildNet~\cite{lee2022wildnet} & RN101 & 47.01 & 50.94 & 48.98 \\
SHADE~\cite{zhao2022style} & RN101 & 50.95 & 60.67 & 55.81 \\ 
\midrule
HGFormer~\cite{ding2023hgformer} & Swin-L & 61.50 & 72.10 & 66.80 \\
VLTSeg~\cite{hummer2023vltseg} & EVA02-L & 64.40 & 76.40 & 70.40 \\
Rein~\cite{wei2024stronger}  &  EVA02-L & 64.10 & 69.50 & 66.80 \\
Rein~\cite{wei2024stronger} & DINOv2-L & 65.00 & 72.30 & 68.65 \\
SET~\cite{yi2024learning} & DINOv2-L & 65.07 & 75.67 & 70.37 \\
tqdm~\cite{pak2025textual}  & EVA02-L & 64.72 & 76.15 & 70.44 \\ 
\rowcolor{gray} MFuser & SIGLIP-L & 65.44 & \underline{77.97} & 71.71 \\
\rowcolor{gray}
MFuser & CLIP-L & \underline{65.58} & \textbf{78.10} & \underline{71.84} \\
\rowcolor{gray}
MFuser & EVA02-L & \textbf{65.81} & 77.93 & \textbf{71.87} \\
\bottomrule
\end{tabular}}
\label{tab:main_r2r}
}
\end{table}

\begin{figure*}[t]
	\centering
	\includegraphics[width=0.95\linewidth]{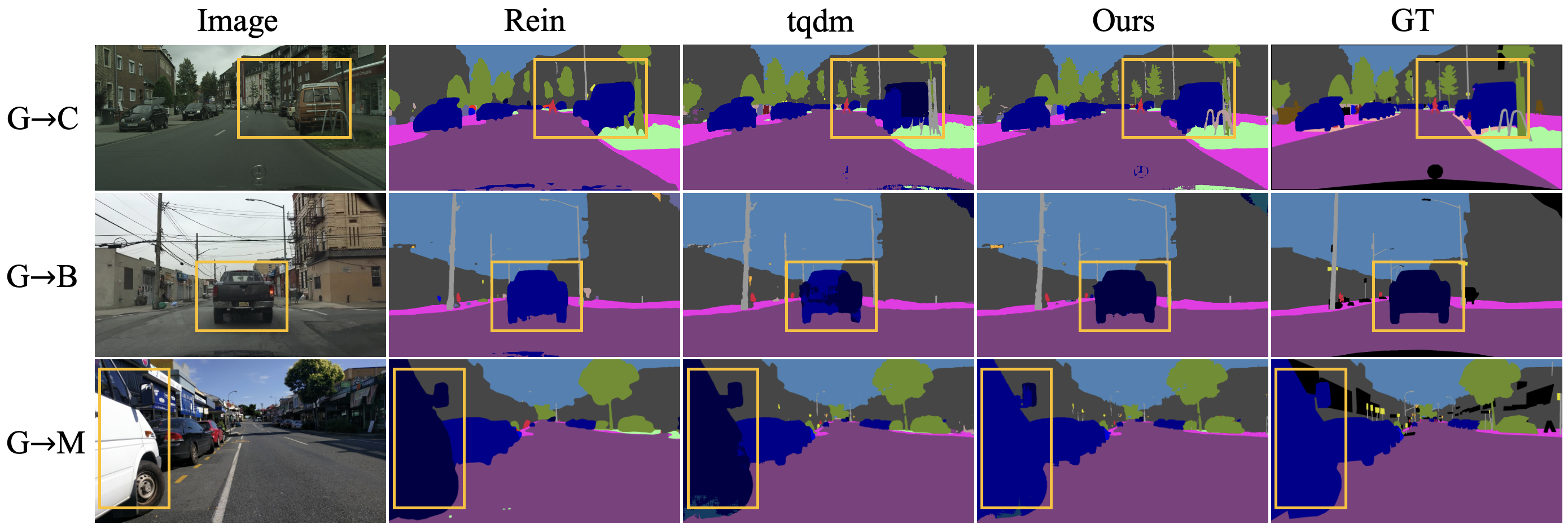}
	\scriptsize%
\setlength\tabcolsep{1pt}%
{%
\newcolumntype{P}[1]{>{\centering\arraybackslash}p{#1}}
\resizebox{0.9\linewidth}{!}{
\begin{tabular}{@{}*{20}{P{0.09\columnwidth}}@{}}
     {\cellcolor[rgb]{0.5,0.25,0.5}}\textcolor{white}{road} 
     &{\cellcolor[rgb]{0.957,0.137,0.91}}sidew. 
     &{\cellcolor[rgb]{0.275,0.275,0.275}}\textcolor{white}{build.} 
     &{\cellcolor[rgb]{0.4,0.4,0.612}}\textcolor{white}{wall} 
     &{\cellcolor[rgb]{0.745,0.6,0.6}}fence 
     &{\cellcolor[rgb]{0.6,0.6,0.6}}pole 
     &{\cellcolor[rgb]{0.98,0.667,0.118}}tr. light
     &{\cellcolor[rgb]{0.863,0.863,0}}tr. sign 
     &{\cellcolor[rgb]{0.42,0.557,0.137}}veget. 
     &{\cellcolor[rgb]{0.596,0.984,0.596}}terrain 
     &{\cellcolor[rgb]{0.275,0.510,0.706}}sky
     &{\cellcolor[rgb]{0.863,0.078,0.235}}\textcolor{white}{person} 
     &{\cellcolor[rgb]{0.988,0.494,0.635}}\textcolor{black}{rider} 
     &{\cellcolor[rgb]{0,0,0.557}}\textcolor{white}{car} 
     &{\cellcolor[rgb]{0,0,0.275}}\textcolor{white}{truck} 
     &{\cellcolor[rgb]{0,0.235,0.392}}\textcolor{white}{bus}
     &{\cellcolor[rgb]{0,0.392,0.471}}\textcolor{white}{train} 
     &{\cellcolor[rgb]{0,0,0.902}}\textcolor{white}{m.bike} 
     & {\cellcolor[rgb]{0.467,0.043,0.125}}\textcolor{white}{bike}
     &{\cellcolor[rgb]{0,0,0}}\textcolor{white}{n/a.}
\end{tabular}}
}%

	\caption{Qualitative results on unseen target domains under the G$\rightarrow$\{C, B, M\} setting. MFuser is compared with Rein~\cite{wei2024stronger} and tqdm~\cite{pak2025textual}.}
	\vspace{-2mm}
	\label{fig:quali}
\end{figure*}

\subsection{In-Depth Analysis}

\begin{table}[h]
	\centering
	\setlength{\tabcolsep}{3pt} 
	\renewcommand{\arraystretch}{0.8} 
	\caption{Efficiency analysis. The experiments are conducted with DINOv2 and EVA02-CLIP models under the G$\rightarrow$\{C, B, M\} settings. The best results are highlighted in \textbf{bold}.}
	\resizebox{\columnwidth}{!}{
		{\begin{tabular}{lcccccc}
				\toprule
				& Params. (M) & FLOPs (G) & C & B & M & Avg. \\
				\midrule
				self-attn (concat.) & 4.20 & 98.64 &  70.24 & 62.31 & 71.11 & 67.89 \\
				self-attn (separate) & 8.40 & 71.08 & 69.68 & 61.91 & 70.85 & 67.48 \\
				bi-deform-attn & 3.35 & 34.65 & 69.46 & 61.17 & 70.11 & 66.91 \\
				\rowcolor{gray}
				MVFuser & 1.67 & 17.21 &  \textbf{70.19} &  \textbf{63.13} &  \textbf{71.28} &  \textbf{68.20} \\
				\bottomrule
		\end{tabular}}
		\label{tab:efficiency}
	}
\end{table}
\paragraph{Efficiency Analysis}
MVFuser is more efficient than \textsl{self}-attention-based adapters, which have quadratic complexity in modeling inter-patch relationships.
To evaluate this, we replace MVFuser with 3 self-attention-based adapters while keeping all other components intact:  $\rm self\mbox{-}attn(concat.)$: $\rm attn(q, k, v\mbox{=}\rm concat(F_{\rm VFM}, F_{\rm VLM}))$; $\rm self\mbox{-}attn(separate)$:  $\{\rm attn(q\mbox{=}F_{\rm VFM}, k,v\mbox{=}F_{\rm VLM}), \rm attn(q\mbox{=}F_{\rm VLM}, k,v\mbox{=}F_{\rm VFM})\}$. $\rm bi\mbox{-}deform\mbox{-}attn$ applies $\rm self\mbox{-}attn(concat.)$ using bidirectional deformable self-attention from Deformable DETR~\cite{zhu2020deformable}. Tab.~\ref{tab:efficiency} summarizes efficiency and results, with parameters and \textsl{FLOPs} per adapter (using $\rm DeepSpeed$ package, $\rm batch \ size$=2). MVFuser achieves the best while significantly reducing parameters and \textsl{FLOPs}.

\vspace{-2mm}
\paragraph{Foundation Model Ensemble}
It is natural to consider ensembling multiple foundation models to enhance performance. To rigorously assess the effectiveness of the proposed MFuser, we address the following questions: \textsl{1) Is simply combining multi-encoder features sufficient to achieve the desired results? 2) Can any parameter-efficient fine-tuning method alone achieve comparable results?}

To answer the first question, we replaced the MVFuser with a simple concatenation of features from the VFM and VLM visual encoders. We also evaluated using only the VFM or VLM visual features independently. As shown in Tab.~\ref{tab:ablate_fusion}, merely concatenating the features from both encoders does not yield satisfactory results and even performs worse than using only VFM or VLM features alone. This occurs because the frozen VFM features are not aligned with the text queries when both are input into the decoder. Additionally, the alignment between VLM visual features and text queries is compromised when the VLM features are mixed with the VFM features.

\begin{table}[t]
\centering
\caption{Ablation studies on the vision feature fusion under the G$\rightarrow$\{C, B, M\} setting. DINOv2 and EVA02-CLIP are applied as the VFM and the VLM, respectively. w.o finetune: directly concatenate features of the two encoders; Conv: utilize convolution layers for fusion; Cross-Attention: implement cross-attention in~\cite{wei2024stronger} for fusion. The best results are highlighted in \textbf{bold}.}
\resizebox{0.8\columnwidth}{!}{
{\begin{tabular}{l|c|c|c|c}
\toprule
Fusion Choice   &  C & B & M & Avg.\\ 
\midrule
VFM-only & 67.68 & 60.82 & 66.89 & 65.13 \\
VLM-only & 68.26 & 60.02 & 70.18 & 66.15 \\
w.o Fintune & 66.96 & 58.88 & 68.25& 64.70\\
Convolution & 69.28 & 61.45 & 69.78 & 66.83 \\
Cross-Attention & 69.67 & 60.52 & 70.43 & 66.87 \\
Sep. MVFuser & 69.57 & 62.88 & 70.59 & 67.68  \\
MVFuser & \textbf{70.19} & \textbf{63.13} & \textbf{71.28} & \textbf{68.20} \\
\bottomrule
\end{tabular}}
\label{tab:ablate_fusion}
}
\end{table}

\begin{figure}[t]
	\centering
	\includegraphics[width=1\linewidth]{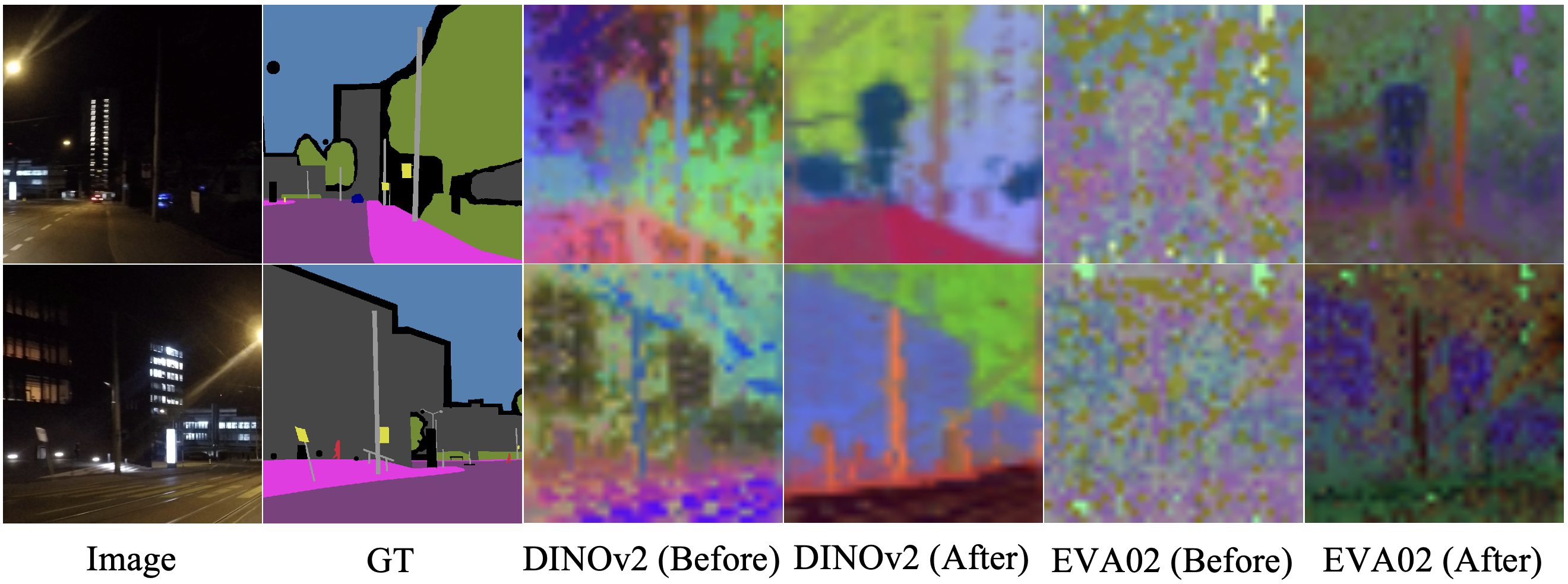}
	\caption{PCA visualization of features from DINOv2 and EVA02-CLIP, illustrating how MVFuser-based adaptation refines their distributions before and after tuning.}
	\label{fig:map}
	\vspace{-2mm}
\end{figure}

Furthermore, fully fine-tuning both encoders is challenging. For example, fully fine-tuning the EVA02-CLIP visual encoder alone requires 4$\times$80GB A100 GPUs for 20 hours, as reported in~\cite{pak2025textual}, which imposes a significant computational burden—let alone the cost of fine-tuning two encoders simultaneously. Alternatively, our MFuser keeps the original VFM and VLM parameters fixed and introduces an additional fusion block, MVFuser, which acts as a bridge between the two foundation models. By optimizing only the MVFuser, we not only adapt the features of both encoders to be more effective but also facilitate interactions between them. Consequently, our method provides a more efficient and effective approach for promoting DGSS with foundation models, achieving the best performance with only 15 hours of training on a single 24GB GPU. Fig.~\ref{fig:map} shows that our proposed MVFuser significantly improves the localization and robustness of the features.

To answer the second question, we implement two alternative adapters to fine-tune the two encoders, based on convolution and attention mechanisms, respectively. For the convolution-based adapter, we first reshape the 1D patch sequence into a 2D feature map and then employ an architecture similar to the spatial branch of the MVFuser, replacing 1D convolutions with 2D convolutions. The attention-based adapter reimplements Rein~\cite{wei2024stronger} to jointly fine-tune both encoders using a single set of learnable tokens through cross-attention. We do not include a self-attention-based adapter due to its quadratic computational cost with respect to the number of tokens, which makes it impractical. As shown in Table \ref{tab:ablate_fusion}, our Mamba-based MVFuser significantly outperforms both the convolution-based and attention-based adapters. This is understandable, as the convolution-based adapter captures only local information, while cross-attention struggles to model token dependencies. Conversely, the Mamba-based MVFuser efficiently captures sequential dynamics with linear complexity. 

In our implementation of MVFuser, VFM features are concatenated before VLM visual features, aiming to enhance VLM features through Mamba's sequential modeling. To evaluate this, we implemented separate MVFuser for DINOv2 and EVA02-CLIP, disentangling their connection. It can be observed from Tab.~\ref{tab:ablate_fusion} that this leads to performance drops, demonstrating the effectiveness of feature interaction. We provide more insights into MVFuser's effectiveness in the supplement. 

\paragraph{Foundation Model Choices}
It remains uncertain whether the performance gain arises from the complementary effects between the VFM and the VLM, or if any two foundation models could achieve similar results. Our method is based on the premise that, while both VFMs and VLMs demonstrate strong robustness, they possess distinct properties due to their different training principles. Consequently, MFuser leverages these differences to complementarily enhance the model's generalization capabilities. 

To verify this, we conduct experiments using two VLMs, where the additional VLM serves as the VFM by utilizing only its visual encoder. Two combinations are tested: ``SIGLIP + EVA02-CLIP'' and ``CLIP + EVA02-CLIP'' with EVA02-CLIP functioning as the VLM while SIGLIP or CLIP acts as the VFM. Evaluation is conducted under the G$\rightarrow$\{C, B, M\} setting, and results are presented in Tab. \ref{tab:ablate_fm_choices}. Both combinations show slight performance improvements over the ``VLM-only'' in Tab. \ref{tab:ablate_fusion}, yet they fall significantly short of any ``VFM + VLM'' pairing. This suggests that the complementary effects between VFMs and VLMs are much more significant than those observed among VLMs alone. Additional evaluations on other VFMs beyond DINOv2 are provided in the supplement.
\begin{table}[t]
\centering
\caption{Ablation studies on the used foundation models. VFM + VLM: only the visual encoder is used when a VLM serve as the VFM. The experiments are conducted under the G$\rightarrow$\{C, B, M\} setting. The best results are highlighted in \textbf{bold}.}
\resizebox{0.8\columnwidth}{!}{
{\begin{tabular}{l|c|c|c|c}
\toprule
VFM + VLM  &  C & B & M & Avg.\\ 
\midrule
SIGLIP + EVA02  & 68.48 & 60.98 & 69.26 & 66.24 \\
CLIP + EVA02 & 68.78 & 61.17 & 70.21 & 66.72 \\
DINOv2 + CLIP & \textbf{71.24} & 61.08 & 71.14 & 67.82 \\
DINOv2 + SIGLIP & 71.10 & 61.19 & \textbf{71.71} & 68.00 \\
DINOv2 + EVA02 & 70.19 & \textbf{63.13} & 71.28 & \textbf{68.20}  \\
\bottomrule
\end{tabular}}
\label{tab:ablate_fm_choices}
}
\end{table}

\paragraph{Text Queries Enhancement}
Solely using class names to obtain text embeddings for each class may not adequately adapt to diverse image types. Encoding image-specific information with text embeddings has been a common practice. In this section, we evaluate the effectiveness of the proposed MTEnhancer under the ``G$\rightarrow$\{C, B, M\}'' setting using DINOv2 and EVA02-CLIP. As demonstrated in Tab.~\ref{tab:ablate_text}, the advantages provided by MTEnhancer are evident. Notably, the hybrid architecture that incorporates self-attention with the conditional Mamba proves to be effective. Furthermore, MTEnhancer outperforms the approach of utilizing cross-attention to encode visual priors. 

\begin{table}[t]
\centering
\caption{Ablation studies on the text embeddings enhancement. Experiments use DINOv2 and EVA02-CLIP under the G$\rightarrow$\{C, B, M\} settings. The best results are highlighted in \textbf{bold}.}
\resizebox{0.8\columnwidth}{!}{
\begin{tabular}{l|c|c|c|c}
	\toprule
	Enhancement Choice & C    & B     & M     & Avg.  \\
	 \midrule
	 w.o. Enhance & 69.57 & 60.83 & 70.32 & 66.91 \\
	 w.o. Hybrid & 69.62 & 61.90 & 70.67 & 67.40 \\
	 Cross-Attention & 69.88 & 61.26 & 70.78 & 67.31 \\
	 MTEnhancer & \textbf{70.19} & \textbf{63.13} & \textbf{71.28} & \textbf{68.20} \\
	\bottomrule
\end{tabular}
\label{tab:ablate_text}
}
\end{table}

\section{Conclusions}
In this work, we proposed MFuser, a novel fusion framework designed to integrate VFMs and VLMs for DGSS. By leveraging the complementary strengths of VFMs and VLMs, MFuser addresses the challenges of increased patch tokens through efficient, scalable fusion with linear complexity. The framework incorporates two key components: MVFuser, which jointly fine-tunes VFMs and VLMs to enhance feature interaction, and MTEnhancer, which refines text embeddings using image priors for better alignment and robustness. Extensive experimental results demonstrate that MFuser achieves precise feature localization and robust text alignment while outperforming state-of-the-art DGSS methods across various benchmarks. The study underscores the potential of combining VFMs and VLMs to achieve superior generalization capabilities in semantic segmentation tasks, and highlights MFuser's effectiveness in advancing DGSS by improving generalization to unseen domains without adding significant computational overhead.

 \clearpage
\setcounter{page}{1}
\maketitlesupplementary

\section{Evaluate on Additional VFMs}
Besides DINOv2 in the main text, we additionally evaluate VFMs, BEiT2~\cite{peng2022beit} and iBOT~\cite{zhou2021ibot}. Both of them are of the \textsl{Large} size. EVA02-CLIP is utilized as the VLM. As shown in Tab.~\ref{tab:vfms}, they also improve the performance of solely using VLM.

\begin{table}[h]
	\centering
	\caption{Ablation studies on more VFMs under the G$\rightarrow$\{C, B, M\} setting. EVA02-CLIP is utilized as the VLM by default. BEiT2~\cite{peng2022beit} and iBOT~\cite{zhou2021ibot} are evaluated as VFMs, respectively. Both are of \textsl{Large} types. }
	\resizebox{0.7\columnwidth}{!}{
		{\begin{tabular}{l|c|c|c|c}
				\toprule
				   &  C & B & M & Avg.\\ 
				\midrule
				VLM-only & 68.26 & 60.02 & 70.18 & 66.15 \\
				+ BEiT2-L & 69.60 & 60.19 & 70.39 & 66.73 \\
				+ iBOT-L & 69.37 & 60.76 & 70.53 & 66.89 \\
				\bottomrule
		\end{tabular}}
		\label{tab:vfms}
	}
\vspace{-4mm}
\end{table}

\section{Evaluate on SYNTHIA Benchmarks}
We compare the performance of the proposed MFuser with existing state-of-the-art DGSS methods under the Synthia$\rightarrow$\{C, B, M\} (as shown in Tab.~\ref{tab:supp_syn}), G$\rightarrow$Synthia and C$\rightarrow$Synthia (as shown in Tab.~\ref{tab:gc2s}) settings. MFuser achieves the best performance on all settings.

\begin{table}
	\centering
	\caption{
	Performance comparison (mIoU in \%) under the synthetic-to-real setting (S$\rightarrow$$\{$C, B, M$\}$). Note that we implement DINOv2~\cite{oquab2023dinov2} as the VFM and EVA02-CLIP~\cite{fang2024eva} as the VLM. Our method is marked in \colorbox{gray}{gray}. The best and second-best results are highlighted in \textbf{bold} and \underline{underlined}, respectively.}
	\resizebox{1\columnwidth}{!}{
	{\begin{tabular}{l|l|c|c|c|c}
			\toprule
			\multirow{2}{4em}{Method} & \multirow{2}{3em}{Backbone} & \multicolumn{4}{c}{\textmd{\textbf{synthetic-to-real}}} \\
			&   & S$\rightarrow$C & S$\rightarrow$B & S$\rightarrow$M & Avg. \\ 
			\midrule
			SAN-SAW~\cite{peng2022semantic} & RN101 & 40.87 & 35.98 & 37.26 & 38.04 \\ 
			TLDR~\cite{kim2023texture} & RN101  & 42.60 &  35.46 & 37.46 & 38.51 \\  
			IBAFormer~\cite{sun2023ibaformer} & MiT-B5 & \underline{50.92} & 44.66 & \underline{50.58} & \underline{48.72} \\ 
			Rein~\cite{wei2024stronger} & DINOv2-L & 48.59 & 44.42 & 48.64 & 47.22 \\
			SET~\cite{yi2024learning} & DINOv2-L & 49.65 & \underline{45.45} & 49.45 & 48.18 \\
			\rowcolor{gray}
			MFuser & EVA02-L & \textbf{54.17} & \textbf{46.67} & \textbf{53.22} & \textbf{51.35} \\
			\bottomrule
	\end{tabular}}
	\label{tab:supp_syn}}
\end{table}

\begin{table}[h]
	\centering
	\caption{
	Performance comparison (mIoU in \%) under G$\rightarrow$S and C$\rightarrow$S. Note that we implement DINOv2~\cite{oquab2023dinov2} as the VFM and EVA02-CLIP~\cite{fang2024eva} as the VLM. Our method is marked in \colorbox{gray}{gray}. The best and second-best results are highlighted in \textbf{bold} and \underline{underlined}, respectively.}
	\resizebox{0.9\columnwidth}{!}{
		{\begin{tabular}{l|c|c|c}
				\toprule
				Method & Backbone & G$\rightarrow$Synthia & C$\rightarrow$Synthia \\
				\midrule
				Rein~\cite{wei2024stronger} & DINOv2-L & 48.86 & 48.56 \\ 
				SET~\cite{yi2024learning} & DINOv2-L & 50.01 & 49.61 \\
				tqdm~\cite{pak2025textual} & EVA02-L & \underline{53.32} & \underline{50.62} \\
				\rowcolor{gray}
				MFuser & EVA02-L & \textbf{54.04} & \textbf{54.13} \\
				\bottomrule
		\end{tabular}}
		\label{tab:gc2s}
	}
\end{table}

\section{Evaluate on ACDC Benchmarks}
We compare the performance of the proposed MFuser with existing state-of-the-art DGSS methods under the clear-to-adverse-weather setting. Models are trained on Cityscapes and tested on ACDC which is composed of four domains, namely \textsl{fog}, \textsl{night}, \textsl{rain} and \textsl{snow}. As shown in Tab.~\ref{tab: acdc}, we consistently outperform the existing methods by a large margin. Particularly, we surpass SET on \textsl{rain} by 3.79 mIoU.

\begin{table}[t]
	\centering
	\caption{Performance comparison (mIoU in \%) on Cityscapes$\rightarrow$ACDC. Note that we implement DINOv2~\cite{oquab2023dinov2} as the VFM and EVA02-CLIP~\cite{fang2024eva} as the VLM. Our method is marked in \colorbox{gray}{gray}. The best and second-best results are highlighted in \textbf{bold} and \underline{underlined}, respectively.}
	\resizebox{1\columnwidth}{!}{
		{\begin{tabular}{l|l|c|c|c|c|c}
				\toprule
				\multirow{2}{4em}{Method} & \multirow{2}{3em}{Backbone} & \multicolumn{5}{c}{\textmd{\textbf{clear-to-adverse-weather}}} \\
				&   & $\rightarrow$Fog & $\rightarrow$Night & $\rightarrow$Rain & $\rightarrow$Snow & Avg. \\ 
				\midrule
				IBN~\cite{pan2018two} & RN50 & 63.80 & 21.20 & 50.40 & 49.60 & 46.25 \\
				IW~\cite{pan2019switchable} & RN50 & 62.40 &	21.80 &	52.40 &	47.60 &	46.05 \\
				ISW~\cite{choi2021robustnet} & RN50 & 64.30 &	24.30 &	56.00 &	49.80 &	48.60 \\
				ISSA~\cite{li2023intra} & MiT-B5 & 67.50 & 33.20 & 55.90 & 53.20 & 52.45 \\
				CMFormer~\cite{bi2024learning} & Swin-L & 77.80 & 33.70 & 67.60 & 64.30 & 60.85 \\
				Rein~\cite{wei2024stronger} & DINOv2-L & 79.48	& 55.92	& 72.45	& 70.57 & 69.61 \\
				SET~\cite{yi2024learning} & DINOv2-L & 80.06 & \underline{57.29} & \underline{74.80} & \underline{73.69} & \underline{71.46} \\
				tqdm~\cite{pak2025textual} & EVA02-L & \underline{81.28}	& 54.80	& 72.92	& 72.41	& 70.35 \\
				\rowcolor{gray}
				MFuser & EVA02-L & \textbf{82.33} & \textbf{57.94} & \textbf{78.59} & \textbf{74.93} & \textbf{73.45} \\
				\bottomrule
		\end{tabular}}
		\label{tab: acdc}
	}
\end{table}

\section{Ablation on the Number of MVFusers}
We evaluate the effect of the number of MVFusers utilized for feature fusion. To do so, MVFuser is inserted after every $N$ blocks. As shown in Tab.~\ref{tab:num_mvfusers}, more MVFusers generally improve performance.

\begin{table}[h]
	\centering
	\caption{Ablation studies on the number of MVFusers under the G$\rightarrow$\{C, B, M\} setting. Note that we implement DINOv2~\cite{oquab2023dinov2} as the VFM and EVA02-CLIP~\cite{fang2024eva} as the VLM. }
	\resizebox{0.6\columnwidth}{!}{
		{\begin{tabular}{l|c|c|c|c}
				\toprule
				$N$ &  C & B & M & Avg.\\ 
				\midrule
				8 & 69.20 & 61.85 & 69.24 & 66.76 \\
				4 & 68.02 & 61.69 & 69.96 & 66.56 \\
				2 & 70.49 & 62.71 & 70.78 & 67.99 \\
				1 & \textbf{70.19} & \textbf{63.13} & \textbf{71.28} & \textbf{68.20} \\ 
				\bottomrule
		\end{tabular}}
		\label{tab:num_mvfusers}
	}
\end{table}

\section{More Qualitative Results}

\begin{figure*}[t]
	\centering
	\includegraphics[width=1.0\linewidth]{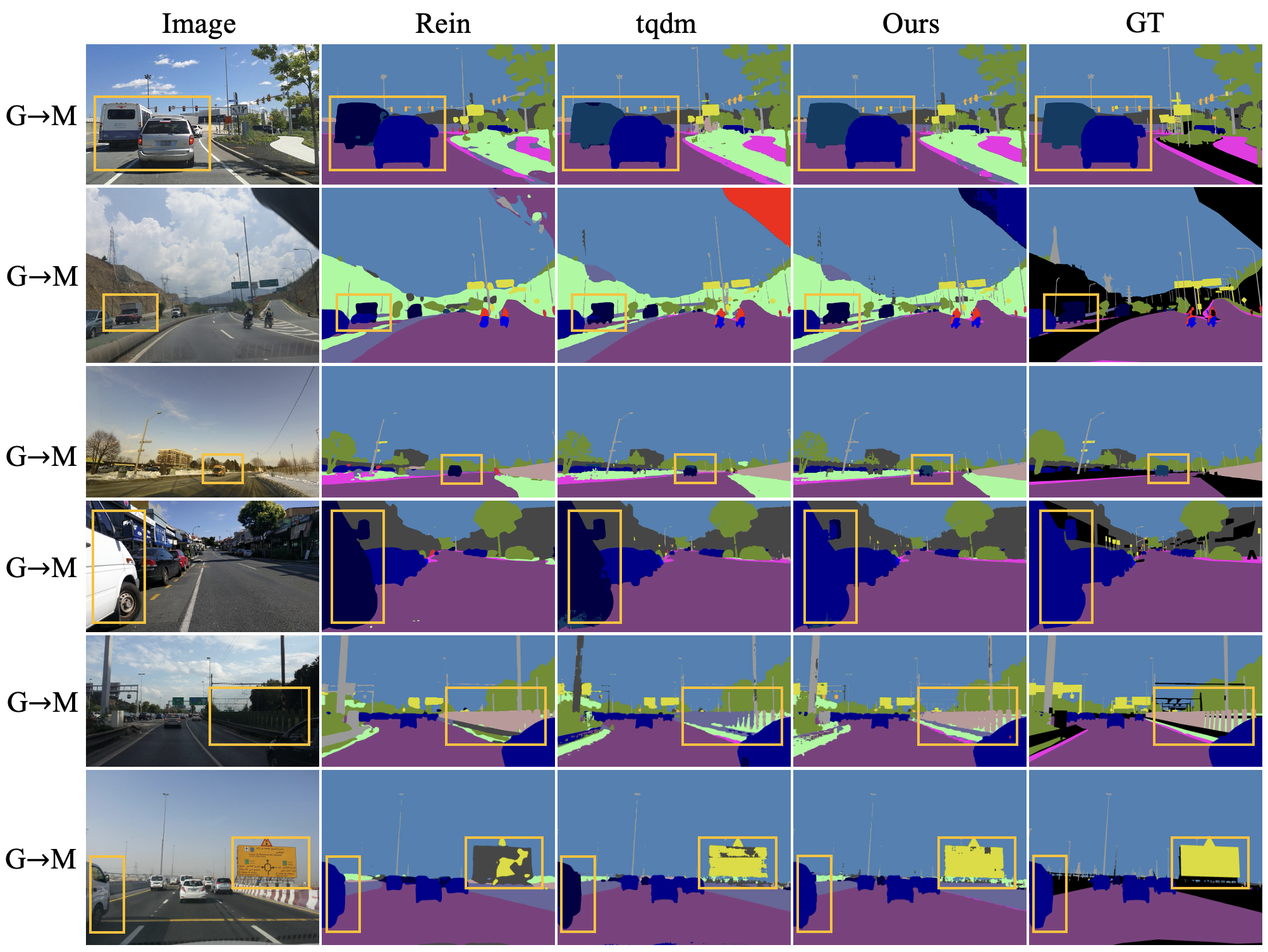}
	\scriptsize%
\setlength\tabcolsep{1pt}%
{%
\newcolumntype{P}[1]{>{\centering\arraybackslash}p{#1}}
\resizebox{0.9\linewidth}{!}{
\begin{tabular}{@{}*{20}{P{0.09\columnwidth}}@{}}
     {\cellcolor[rgb]{0.5,0.25,0.5}}\textcolor{white}{road} 
     &{\cellcolor[rgb]{0.957,0.137,0.91}}sidew. 
     &{\cellcolor[rgb]{0.275,0.275,0.275}}\textcolor{white}{build.} 
     &{\cellcolor[rgb]{0.4,0.4,0.612}}\textcolor{white}{wall} 
     &{\cellcolor[rgb]{0.745,0.6,0.6}}fence 
     &{\cellcolor[rgb]{0.6,0.6,0.6}}pole 
     &{\cellcolor[rgb]{0.98,0.667,0.118}}tr. light
     &{\cellcolor[rgb]{0.863,0.863,0}}tr. sign 
     &{\cellcolor[rgb]{0.42,0.557,0.137}}veget. 
     &{\cellcolor[rgb]{0.596,0.984,0.596}}terrain 
     &{\cellcolor[rgb]{0.275,0.510,0.706}}sky
     &{\cellcolor[rgb]{0.863,0.078,0.235}}\textcolor{white}{person} 
     &{\cellcolor[rgb]{0.988,0.494,0.635}}\textcolor{black}{rider} 
     &{\cellcolor[rgb]{0,0,0.557}}\textcolor{white}{car} 
     &{\cellcolor[rgb]{0,0,0.275}}\textcolor{white}{truck} 
     &{\cellcolor[rgb]{0,0.235,0.392}}\textcolor{white}{bus}
     &{\cellcolor[rgb]{0,0.392,0.471}}\textcolor{white}{train} 
     &{\cellcolor[rgb]{0,0,0.902}}\textcolor{white}{m.bike} 
     & {\cellcolor[rgb]{0.467,0.043,0.125}}\textcolor{white}{bike}
     &{\cellcolor[rgb]{0,0,0}}\textcolor{white}{n/a.}
\end{tabular}}
}%

	\caption{Qualitative results on unseen target domains under the G$\rightarrow$M setting. MFuser is compared with Rein~\cite{wei2024stronger} and tqdm~\cite{pak2025textual}.}
	\label{fig:quali}
\end{figure*}

\begin{figure*}[t]
	\centering
	\includegraphics[width=1.0\linewidth]{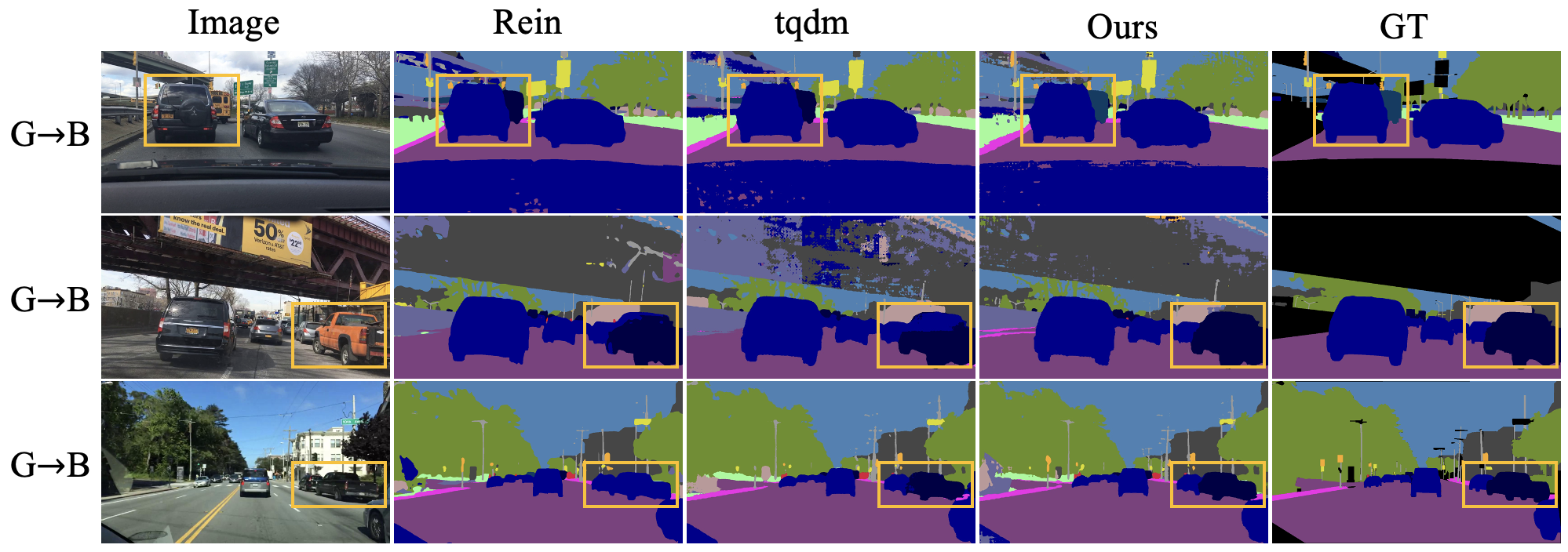}
	\scriptsize%
\setlength\tabcolsep{1pt}%
{%
\newcolumntype{P}[1]{>{\centering\arraybackslash}p{#1}}
\resizebox{0.9\linewidth}{!}{
\begin{tabular}{@{}*{20}{P{0.09\columnwidth}}@{}}
     {\cellcolor[rgb]{0.5,0.25,0.5}}\textcolor{white}{road} 
     &{\cellcolor[rgb]{0.957,0.137,0.91}}sidew. 
     &{\cellcolor[rgb]{0.275,0.275,0.275}}\textcolor{white}{build.} 
     &{\cellcolor[rgb]{0.4,0.4,0.612}}\textcolor{white}{wall} 
     &{\cellcolor[rgb]{0.745,0.6,0.6}}fence 
     &{\cellcolor[rgb]{0.6,0.6,0.6}}pole 
     &{\cellcolor[rgb]{0.98,0.667,0.118}}tr. light
     &{\cellcolor[rgb]{0.863,0.863,0}}tr. sign 
     &{\cellcolor[rgb]{0.42,0.557,0.137}}veget. 
     &{\cellcolor[rgb]{0.596,0.984,0.596}}terrain 
     &{\cellcolor[rgb]{0.275,0.510,0.706}}sky
     &{\cellcolor[rgb]{0.863,0.078,0.235}}\textcolor{white}{person} 
     &{\cellcolor[rgb]{0.988,0.494,0.635}}\textcolor{black}{rider} 
     &{\cellcolor[rgb]{0,0,0.557}}\textcolor{white}{car} 
     &{\cellcolor[rgb]{0,0,0.275}}\textcolor{white}{truck} 
     &{\cellcolor[rgb]{0,0.235,0.392}}\textcolor{white}{bus}
     &{\cellcolor[rgb]{0,0.392,0.471}}\textcolor{white}{train} 
     &{\cellcolor[rgb]{0,0,0.902}}\textcolor{white}{m.bike} 
     & {\cellcolor[rgb]{0.467,0.043,0.125}}\textcolor{white}{bike}
     &{\cellcolor[rgb]{0,0,0}}\textcolor{white}{n/a.}
\end{tabular}}
}%

	\caption{Qualitative results on unseen target domains under the G$\rightarrow$B setting. MFuser is compared with Rein~\cite{wei2024stronger} and tqdm~\cite{pak2025textual}.}
	\label{fig:quali}
\end{figure*}

\clearpage
{
    \small
    \bibliographystyle{ieeenat_fullname}
    \bibliography{main}

\begin{thebibliography}{72}
\providecommand{\natexlab}[1]{#1}
\providecommand{\url}[1]{\texttt{#1}}
\expandafter\ifx\csname urlstyle\endcsname\relax
  \providecommand{\doi}[1]{doi: #1}\else
  \providecommand{\doi}{doi: \begingroup \urlstyle{rm}\Url}\fi

\bibitem[Ai et~al.(2024)Ai, Qi, Wang, Cheng, Wang, and Tan]{ai2024domain}
Yihao Ai, Yifei Qi, Bo Wang, Yu Cheng, Xinchao Wang, and Robby~T Tan.
\newblock Domain-adaptive 2d human pose estimation via dual teachers in
  extremely low-light conditions.
\newblock In \emph{European Conference on Computer Vision}, pages 221--239.
  Springer, 2024.

\bibitem[Benigmim et~al.(2024)Benigmim, Roy, Essid, Kalogeiton, and
  Lathuili{\`e}re]{benigmim2024collaborating}
Yasser Benigmim, Subhankar Roy, Slim Essid, Vicky Kalogeiton, and St{\'e}phane
  Lathuili{\`e}re.
\newblock Collaborating foundation models for domain generalized semantic
  segmentation.
\newblock In \emph{Proceedings of the IEEE/CVF Conference on Computer Vision
  and Pattern Recognition}, pages 3108--3119, 2024.

\bibitem[Bi et~al.(2024)Bi, You, and Gevers]{bi2024learning}
Qi Bi, Shaodi You, and Theo Gevers.
\newblock Learning content-enhanced mask transformer for domain generalized
  urban-scene segmentation.
\newblock In \emph{Proceedings of the AAAI Conference on Artificial
  Intelligence}, pages 819--827, 2024.

\bibitem[Caron et~al.(2021)Caron, Touvron, Misra, J{\'e}gou, Mairal,
  Bojanowski, and Joulin]{caron2021emerging}
Mathilde Caron, Hugo Touvron, Ishan Misra, Herv{\'e} J{\'e}gou, Julien Mairal,
  Piotr Bojanowski, and Armand Joulin.
\newblock Emerging properties in self-supervised vision transformers.
\newblock In \emph{Proceedings of the IEEE/CVF international conference on
  computer vision}, pages 9650--9660, 2021.

\bibitem[Chattopadhyay et~al.(2023)Chattopadhyay, Sarangmath, Vijaykumar, and
  Hoffman]{chattopadhyay2023pasta}
Prithvijit Chattopadhyay, Kartik Sarangmath, Vivek Vijaykumar, and Judy
  Hoffman.
\newblock Pasta: Proportional amplitude spectrum training augmentation for
  syn-to-real domain generalization.
\newblock In \emph{Proceedings of the IEEE/CVF International Conference on
  Computer Vision}, pages 19288--19300, 2023.

\bibitem[Chen et~al.(2024)Chen, Lin, Jin, Yan, Ye, Yuan, and Tan]{chen2024dual}
Tingting Chen, Beibei Lin, Yeying Jin, Wending Yan, Wei Ye, Yuan Yuan, and
  Robby~T Tan.
\newblock Dual-rain: Video rain removal using assertive and gentle teachers.
\newblock In \emph{European Conference on Computer Vision}, pages 127--143.
  Springer, 2024.

\bibitem[Chen et~al.(2023)Chen, Qi, Cao, Li, Meng, and Meng]{chen2023class}
Zitan Chen, Zhuang Qi, Xiao Cao, Xiangxian Li, Xiangxu Meng, and Lei Meng.
\newblock Class-level structural relation modeling and smoothing for visual
  representation learning.
\newblock In \emph{Proceedings of the 31st ACM International Conference on
  Multimedia}, pages 2964--2972, 2023.

\bibitem[Cheng et~al.(2022)Cheng, Misra, Schwing, Kirillov, and
  Girdhar]{cheng2022masked}
Bowen Cheng, Ishan Misra, Alexander~G Schwing, Alexander Kirillov, and Rohit
  Girdhar.
\newblock Masked-attention mask transformer for universal image segmentation.
\newblock In \emph{Proceedings of the IEEE/CVF conference on computer vision
  and pattern recognition}, pages 1290--1299, 2022.

\bibitem[Cho et~al.(2023)Cho, Nam, Kim, Yang, and Kwak]{cho2023promptstyler}
Junhyeong Cho, Gilhyun Nam, Sungyeon Kim, Hunmin Yang, and Suha Kwak.
\newblock Promptstyler: Prompt-driven style generation for source-free domain
  generalization.
\newblock In \emph{Proceedings of the IEEE/CVF International Conference on
  Computer Vision}, pages 15702--15712, 2023.

\bibitem[Choi et~al.(2021)Choi, Jung, Yun, Kim, Kim, and
  Choo]{choi2021robustnet}
Sungha Choi, Sanghun Jung, Huiwon Yun, Joanne~T Kim, Seungryong Kim, and Jaegul
  Choo.
\newblock Robustnet: Improving domain generalization in urban-scene
  segmentation via instance selective whitening.
\newblock In \emph{Proceedings of the IEEE/CVF conference on computer vision
  and pattern recognition}, pages 11580--11590, 2021.

\bibitem[Cordts et~al.(2016)Cordts, Omran, Ramos, Rehfeld, Enzweiler, Benenson,
  Franke, Roth, and Schiele]{cordts2016cityscapes}
Marius Cordts, Mohamed Omran, Sebastian Ramos, Timo Rehfeld, Markus Enzweiler,
  Rodrigo Benenson, Uwe Franke, Stefan Roth, and Bernt Schiele.
\newblock The cityscapes dataset for semantic urban scene understanding.
\newblock In \emph{Proceedings of the IEEE conference on computer vision and
  pattern recognition}, pages 3213--3223, 2016.

\bibitem[Das et~al.(2024)Das, Hu, Jiang, and Schiele]{das2024mta}
Anurag Das, Xinting Hu, Li Jiang, and Bernt Schiele.
\newblock Mta-clip: Language-guided semantic segmentation with mask-text
  alignment.
\newblock In \emph{European Conference on Computer Vision}, pages 39--56.
  Springer, 2024.

\bibitem[Ding et~al.(2023)Ding, Xue, Xia, Schiele, and Dai]{ding2023hgformer}
Jian Ding, Nan Xue, Gui-Song Xia, Bernt Schiele, and Dengxin Dai.
\newblock Hgformer: Hierarchical grouping transformer for domain generalized
  semantic segmentation.
\newblock In \emph{Proceedings of the IEEE/CVF Conference on Computer Vision
  and Pattern Recognition}, pages 15413--15423, 2023.

\bibitem[Fahes et~al.(2023)Fahes, Vu, Bursuc, P{\'e}rez, and
  de~Charette]{fahes2023simple}
Mohammad Fahes, Tuan-Hung Vu, Andrei Bursuc, Patrick P{\'e}rez, and Raoul de
  Charette.
\newblock A simple recipe for language-guided domain generalized segmentation.
\newblock \emph{arXiv preprint arXiv:2311.17922}, 2023.

\bibitem[Fahes et~al.(2024)Fahes, Vu, Bursuc, P{\'e}rez, and
  de~Charette]{fahes2024simple}
Mohammad Fahes, Tuan-Hung Vu, Andrei Bursuc, Patrick P{\'e}rez, and Raoul de
  Charette.
\newblock A simple recipe for language-guided domain generalized segmentation.
\newblock In \emph{Proceedings of the IEEE/CVF Conference on Computer Vision
  and Pattern Recognition}, pages 23428--23437, 2024.

\bibitem[Fang et~al.(2024)Fang, Sun, Wang, Huang, Wang, and Cao]{fang2024eva}
Yuxin Fang, Quan Sun, Xinggang Wang, Tiejun Huang, Xinlong Wang, and Yue Cao.
\newblock Eva-02: A visual representation for neon genesis.
\newblock \emph{Image and Vision Computing}, page 105171, 2024.

\bibitem[Gu and Dao(2023)]{gu2023mamba}
Albert Gu and Tri Dao.
\newblock Mamba: Linear-time sequence modeling with selective state spaces.
\newblock \emph{arXiv preprint arXiv:2312.00752}, 2023.

\bibitem[Gu et~al.(2021)Gu, Goel, and R{\'e}]{gu2021efficiently}
Albert Gu, Karan Goel, and Christopher R{\'e}.
\newblock Efficiently modeling long sequences with structured state spaces.
\newblock \emph{arXiv preprint arXiv:2111.00396}, 2021.

\bibitem[Hatamizadeh and Kautz(2024)]{hatamizadeh2024mambavision}
Ali Hatamizadeh and Jan Kautz.
\newblock Mambavision: A hybrid mamba-transformer vision backbone.
\newblock \emph{arXiv preprint arXiv:2407.08083}, 2024.

\bibitem[He et~al.(2022)He, Chen, Xie, Li, Doll{\'a}r, and
  Girshick]{he2022masked}
Kaiming He, Xinlei Chen, Saining Xie, Yanghao Li, Piotr Doll{\'a}r, and Ross
  Girshick.
\newblock Masked autoencoders are scalable vision learners.
\newblock In \emph{Proceedings of the IEEE/CVF conference on computer vision
  and pattern recognition}, pages 16000--16009, 2022.

\bibitem[Hoyer et~al.(2022)Hoyer, Dai, and Van~Gool]{hoyer2022hrda}
Lukas Hoyer, Dengxin Dai, and Luc Van~Gool.
\newblock Hrda: Context-aware high-resolution domain-adaptive semantic
  segmentation.
\newblock In \emph{European Conference on Computer Vision}, pages 372--391.
  Springer, 2022.

\bibitem[Huang et~al.(2019)Huang, Zhou, Zhu, Liu, and Shao]{huang2019iterative}
Lei Huang, Yi Zhou, Fan Zhu, Li Liu, and Ling Shao.
\newblock Iterative normalization: Beyond standardization towards efficient
  whitening.
\newblock In \emph{Proceedings of the IEEE/CVF conference on computer vision
  and pattern recognition}, pages 4874--4883, 2019.

\bibitem[Huang et~al.(2023{\natexlab{a}})Huang, Chen, Li, Li, Li, Song, Yan,
  and Xiong]{huang2023style}
Wei Huang, Chang Chen, Yong Li, Jiacheng Li, Cheng Li, Fenglong Song, Youliang
  Yan, and Zhiwei Xiong.
\newblock Style projected clustering for domain generalized semantic
  segmentation.
\newblock In \emph{Proceedings of the IEEE/CVF conference on computer vision
  and pattern recognition}, pages 3061--3071, 2023{\natexlab{a}}.

\bibitem[Huang et~al.(2023{\natexlab{b}})Huang, Zhou, Ling, Cai, Wang, and
  Lee]{huang2023sentence}
Zeyi Huang, Andy Zhou, Zijian Ling, Mu Cai, Haohan Wang, and Yong~Jae Lee.
\newblock A sentence speaks a thousand images: Domain generalization through
  distilling clip with language guidance.
\newblock In \emph{Proceedings of the IEEE/CVF International Conference on
  Computer Vision}, pages 11685--11695, 2023{\natexlab{b}}.

\bibitem[H{\"u}mmer et~al.(2023)H{\"u}mmer, Schwonberg, Zhong, Cao, Knoll, and
  Gottschalk]{hummer2023vltseg}
Christoph H{\"u}mmer, Manuel Schwonberg, Liangwei Zhong, Hu Cao, Alois Knoll,
  and Hanno Gottschalk.
\newblock Vltseg: Simple transfer of clip-based vision-language representations
  for domain generalized semantic segmentation.
\newblock \emph{arXiv preprint arXiv:2312.02021}, 2023.

\bibitem[Kim et~al.(2022)Kim, Lee, Park, Min, and Sohn]{kim2022pin}
Jin Kim, Jiyoung Lee, Jungin Park, Dongbo Min, and Kwanghoon Sohn.
\newblock Pin the memory: Learning to generalize semantic segmentation.
\newblock In \emph{Proceedings of the IEEE/CVF Conference on Computer Vision
  and Pattern Recognition}, pages 4350--4360, 2022.

\bibitem[Kim et~al.(2023)Kim, Kim, and Kim]{kim2023texture}
Sunghwan Kim, Dae-hwan Kim, and Hoseong Kim.
\newblock Texture learning domain randomization for domain generalized
  segmentation.
\newblock In \emph{Proceedings of the IEEE/CVF International Conference on
  Computer Vision}, pages 677--687, 2023.

\bibitem[Kirillov et~al.(2023)Kirillov, Mintun, Ravi, Mao, Rolland, Gustafson,
  Xiao, Whitehead, Berg, Lo, et~al.]{kirillov2023segment}
Alexander Kirillov, Eric Mintun, Nikhila Ravi, Hanzi Mao, Chloe Rolland, Laura
  Gustafson, Tete Xiao, Spencer Whitehead, Alexander~C Berg, Wan-Yen Lo, et~al.
\newblock Segment anything.
\newblock In \emph{Proceedings of the IEEE/CVF International Conference on
  Computer Vision}, pages 4015--4026, 2023.

\bibitem[Lee et~al.(2022)Lee, Seong, Lee, and Kim]{lee2022wildnet}
Suhyeon Lee, Hongje Seong, Seongwon Lee, and Euntai Kim.
\newblock Wildnet: Learning domain generalized semantic segmentation from the
  wild.
\newblock In \emph{Proceedings of the IEEE/CVF conference on computer vision
  and pattern recognition}, pages 9936--9946, 2022.

\bibitem[Lei et~al.(2024)Lei, Wang, and Tan]{lei2024ez}
Qinqian Lei, Bo Wang, and Robby Tan.
\newblock Ez-hoi: Vlm adaptation via guided prompt learning for zero-shot hoi
  detection.
\newblock \emph{Advances in Neural Information Processing Systems},
  37:\penalty0 55831--55857, 2024.

\bibitem[Li et~al.(2023{\natexlab{a}})Li, Zhang, Sun, Zou, Liu, Yang, Li,
  Zhang, and Gao]{li2023semantic}
Feng Li, Hao Zhang, Peize Sun, Xueyan Zou, Shilong Liu, Jianwei Yang, Chunyuan
  Li, Lei Zhang, and Jianfeng Gao.
\newblock Semantic-sam: Segment and recognize anything at any granularity.
\newblock \emph{arXiv preprint arXiv:2307.04767}, 2023{\natexlab{a}}.

\bibitem[Li et~al.(2023{\natexlab{b}})Li, Zhang, Keuper, and
  Khoreva]{li2023intra}
Yumeng Li, Dan Zhang, Margret Keuper, and Anna Khoreva.
\newblock Intra-source style augmentation for improved domain generalization.
\newblock In \emph{Proceedings of the IEEE/CVF Winter Conference on
  Applications of Computer Vision}, pages 509--519, 2023{\natexlab{b}}.

\bibitem[Lin et~al.(2024{\natexlab{a}})Lin, Jin, Yan, Ye, Yuan, and
  Tan]{lin2024nighthaze}
Beibei Lin, Yeying Jin, Wending Yan, Wei Ye, Yuan Yuan, and Robby~T Tan.
\newblock Nighthaze: Nighttime image dehazing via self-prior learning.
\newblock \emph{arXiv preprint arXiv:2403.07408}, 2024{\natexlab{a}}.

\bibitem[Lin et~al.(2024{\natexlab{b}})Lin, Jin, Yan, Ye, Yuan, Zhang, and
  Tan]{lin2024nightrain}
Beibei Lin, Yeying Jin, Wending Yan, Wei Ye, Yuan Yuan, Shunli Zhang, and
  Robby~T Tan.
\newblock Nightrain: Nighttime video deraining via adaptive-rain-removal and
  adaptive-correction.
\newblock In \emph{Proceedings of the AAAI Conference on Artificial
  Intelligence}, pages 3378--3385, 2024{\natexlab{b}}.

\bibitem[Liu et~al.(2023)Liu, Zeng, Ren, Li, Zhang, Yang, Li, Yang, Su, Zhu,
  et~al.]{liu2023grounding}
Shilong Liu, Zhaoyang Zeng, Tianhe Ren, Feng Li, Hao Zhang, Jie Yang, Chunyuan
  Li, Jianwei Yang, Hang Su, Jun Zhu, et~al.
\newblock Grounding dino: Marrying dino with grounded pre-training for open-set
  object detection.
\newblock \emph{arXiv preprint arXiv:2303.05499}, 2023.

\bibitem[Mehta et~al.(2022)Mehta, Gupta, Cutkosky, and
  Neyshabur]{mehta2022long}
Harsh Mehta, Ankit Gupta, Ashok Cutkosky, and Behnam Neyshabur.
\newblock Long range language modeling via gated state spaces.
\newblock \emph{arXiv preprint arXiv:2206.13947}, 2022.

\bibitem[Neuhold et~al.(2017)Neuhold, Ollmann, Rota~Bulo, and
  Kontschieder]{neuhold2017mapillary}
Gerhard Neuhold, Tobias Ollmann, Samuel Rota~Bulo, and Peter Kontschieder.
\newblock The mapillary vistas dataset for semantic understanding of street
  scenes.
\newblock In \emph{Proceedings of the IEEE international conference on computer
  vision}, pages 4990--4999, 2017.

\bibitem[Oquab et~al.(2023)Oquab, Darcet, Moutakanni, Vo, Szafraniec, Khalidov,
  Fernandez, Haziza, Massa, El-Nouby, et~al.]{oquab2023dinov2}
Maxime Oquab, Timoth{\'e}e Darcet, Th{\'e}o Moutakanni, Huy Vo, Marc
  Szafraniec, Vasil Khalidov, Pierre Fernandez, Daniel Haziza, Francisco Massa,
  Alaaeldin El-Nouby, et~al.
\newblock Dinov2: Learning robust visual features without supervision.
\newblock \emph{arXiv preprint arXiv:2304.07193}, 2023.

\bibitem[Pak et~al.(2024)Pak, Woo, Kim, Kim, and Kim]{pak2024textual}
Byeonghyun Pak, Byeongju Woo, Sunghwan Kim, Dae-hwan Kim, and Hoseong Kim.
\newblock Textual query-driven mask transformer for domain generalized
  segmentation.
\newblock \emph{arXiv preprint arXiv:2407.09033}, 2024.

\bibitem[Pak et~al.(2025)Pak, Woo, Kim, Kim, and Kim]{pak2025textual}
Byeonghyun Pak, Byeongju Woo, Sunghwan Kim, Dae-hwan Kim, and Hoseong Kim.
\newblock Textual query-driven mask transformer for domain generalized
  segmentation.
\newblock In \emph{European Conference on Computer Vision}, pages 37--54.
  Springer, 2025.

\bibitem[Pan et~al.(2018)Pan, Luo, Shi, and Tang]{pan2018two}
Xingang Pan, Ping Luo, Jianping Shi, and Xiaoou Tang.
\newblock Two at once: Enhancing learning and generalization capacities via
  ibn-net.
\newblock In \emph{Proceedings of the european conference on computer vision
  (ECCV)}, pages 464--479, 2018.

\bibitem[Pan et~al.(2019)Pan, Zhan, Shi, Tang, and Luo]{pan2019switchable}
Xingang Pan, Xiaohang Zhan, Jianping Shi, Xiaoou Tang, and Ping Luo.
\newblock Switchable whitening for deep representation learning.
\newblock In \emph{Proceedings of the IEEE/CVF international conference on
  computer vision}, pages 1863--1871, 2019.

\bibitem[Peng et~al.(2022{\natexlab{a}})Peng, Lei, Hayat, Guo, and
  Li]{peng2022semantic}
Duo Peng, Yinjie Lei, Munawar Hayat, Yulan Guo, and Wen Li.
\newblock Semantic-aware domain generalized segmentation.
\newblock In \emph{Proceedings of the IEEE/CVF conference on computer vision
  and pattern recognition}, pages 2594--2605, 2022{\natexlab{a}}.

\bibitem[Peng et~al.(2022{\natexlab{b}})Peng, Dong, Bao, Ye, and
  Wei]{peng2022beit}
Zhiliang Peng, Li Dong, Hangbo Bao, Qixiang Ye, and Furu Wei.
\newblock Beit v2: Masked image modeling with vector-quantized visual
  tokenizers.
\newblock \emph{arXiv preprint arXiv:2208.06366}, 2022{\natexlab{b}}.

\bibitem[Radford et~al.(2021)Radford, Kim, Hallacy, Ramesh, Goh, Agarwal,
  Sastry, Askell, Mishkin, Clark, et~al.]{radford2021learning}
Alec Radford, Jong~Wook Kim, Chris Hallacy, Aditya Ramesh, Gabriel Goh,
  Sandhini Agarwal, Girish Sastry, Amanda Askell, Pamela Mishkin, Jack Clark,
  et~al.
\newblock Learning transferable visual models from natural language
  supervision.
\newblock In \emph{International conference on machine learning}, pages
  8748--8763. PMLR, 2021.

\bibitem[Rao et~al.(2022)Rao, Zhao, Chen, Tang, Zhu, Huang, Zhou, and
  Lu]{rao2022denseclip}
Yongming Rao, Wenliang Zhao, Guangyi Chen, Yansong Tang, Zheng Zhu, Guan Huang,
  Jie Zhou, and Jiwen Lu.
\newblock Denseclip: Language-guided dense prediction with context-aware
  prompting.
\newblock In \emph{Proceedings of the IEEE/CVF conference on computer vision
  and pattern recognition}, pages 18082--18091, 2022.

\bibitem[Richter et~al.(2016)Richter, Vineet, Roth, and
  Koltun]{richter2016playing}
Stephan~R Richter, Vibhav Vineet, Stefan Roth, and Vladlen Koltun.
\newblock Playing for data: Ground truth from computer games.
\newblock In \emph{Computer Vision--ECCV 2016: 14th European Conference,
  Amsterdam, The Netherlands, October 11-14, 2016, Proceedings, Part II 14},
  pages 102--118. Springer, 2016.

\bibitem[Rombach et~al.(2022)Rombach, Blattmann, Lorenz, Esser, and
  Ommer]{rombach2022high}
Robin Rombach, Andreas Blattmann, Dominik Lorenz, Patrick Esser, and Bj{\"o}rn
  Ommer.
\newblock High-resolution image synthesis with latent diffusion models.
\newblock In \emph{Proceedings of the IEEE/CVF conference on computer vision
  and pattern recognition}, pages 10684--10695, 2022.

\bibitem[Ros et~al.(2016)Ros, Sellart, Materzynska, Vazquez, and
  Lopez]{ros2016synthia}
German Ros, Laura Sellart, Joanna Materzynska, David Vazquez, and Antonio~M
  Lopez.
\newblock The synthia dataset: A large collection of synthetic images for
  semantic segmentation of urban scenes.
\newblock In \emph{Proceedings of the IEEE conference on computer vision and
  pattern recognition}, pages 3234--3243, 2016.

\bibitem[Ruan and Xiang(2024)]{ruan2024vm}
Jiacheng Ruan and Suncheng Xiang.
\newblock Vm-unet: Vision mamba unet for medical image segmentation.
\newblock \emph{arXiv preprint arXiv:2402.02491}, 2024.

\bibitem[Sakaridis et~al.(2021)Sakaridis, Dai, and Van~Gool]{sakaridis2021acdc}
Christos Sakaridis, Dengxin Dai, and Luc Van~Gool.
\newblock {ACDC}: The {Adverse} {Conditions} {Dataset} with {Correspondences}
  for semantic driving scene understanding.
\newblock In \emph{ICCV}, 2021.

\bibitem[Smith et~al.(2022)Smith, Warrington, and
  Linderman]{smith2022simplified}
Jimmy~TH Smith, Andrew Warrington, and Scott~W Linderman.
\newblock Simplified state space layers for sequence modeling.
\newblock \emph{arXiv preprint arXiv:2208.04933}, 2022.

\bibitem[Sun et~al.(2023{\natexlab{a}})Sun, Chen, Zheng, Wu, Felsberg, and
  Tang]{sun2023ibaformer}
Qiyu Sun, Huilin Chen, Meng Zheng, Ziyan Wu, Michael Felsberg, and Yang Tang.
\newblock Ibaformer: Intra-batch attention transformer for domain generalized
  semantic segmentation.
\newblock \emph{arXiv preprint arXiv:2309.06282}, 2023{\natexlab{a}}.

\bibitem[Sun et~al.(2023{\natexlab{b}})Sun, Fang, Wu, Wang, and
  Cao]{sun2023eva}
Quan Sun, Yuxin Fang, Ledell Wu, Xinlong Wang, and Yue Cao.
\newblock Eva-clip: Improved training techniques for clip at scale.
\newblock \emph{arXiv preprint arXiv:2303.15389}, 2023{\natexlab{b}}.

\bibitem[Wei et~al.(2024)Wei, Chen, Jin, Ma, Liu, Ling, Wang, Chen, and
  Zheng]{wei2024stronger}
Zhixiang Wei, Lin Chen, Yi Jin, Xiaoxiao Ma, Tianle Liu, Pengyang Ling, Ben
  Wang, Huaian Chen, and Jinjin Zheng.
\newblock Stronger fewer \& superior: Harnessing vision foundation models for
  domain generalized semantic segmentation.
\newblock In \emph{Proceedings of the IEEE/CVF Conference on Computer Vision
  and Pattern Recognition}, pages 28619--28630, 2024.

\bibitem[Wu et~al.(2024)Wu, Cao, Li, Chen, Wang, Meng, and
  Huang]{wu2024towards}
Yihang Wu, Xiao Cao, Kaixin Li, Zitan Chen, Haonan Wang, Lei Meng, and Zhiyong
  Huang.
\newblock Towards better text-to-image generation alignment via attention
  modulation.
\newblock \emph{arXiv preprint arXiv:2404.13899}, 2024.

\bibitem[Xu et~al.(2022)Xu, Yao, Jiang, Jiang, Chu, Han, Zhang, Wang, and
  Tai]{xu2022dirl}
Qi Xu, Liang Yao, Zhengkai Jiang, Guannan Jiang, Wenqing Chu, Wenhui Han, Wei
  Zhang, Chengjie Wang, and Ying Tai.
\newblock Dirl: Domain-invariant representation learning for generalizable
  semantic segmentation.
\newblock In \emph{Proceedings of the AAAI Conference on Artificial
  Intelligence}, pages 2884--2892, 2022.

\bibitem[Yan et~al.(2023)Yan, Tan, Zeng, and Liu]{Yan_2023_ICCV}
Weilong Yan, Robby~T. Tan, Bing Zeng, and Shuaicheng Liu.
\newblock Deep homography mixture for single image rolling shutter correction.
\newblock In \emph{Proceedings of the IEEE/CVF International Conference on
  Computer Vision (ICCV)}, pages 9868--9877, 2023.

\bibitem[Yang et~al.(2022)Yang, Mi, Yuan, Wang, and Tan]{yang2022object}
Xin Yang, Michael~Bi Mi, Yuan Yuan, Xin Wang, and Robby~T Tan.
\newblock Object detection in foggy scenes by embedding depth and
  reconstruction into domain adaptation.
\newblock In \emph{Proceedings of the Asian Conference on Computer Vision},
  pages 1093--1108, 2022.

\bibitem[Yang et~al.(2024)Yang, Yan, Yuan, Mi, and Tan]{yang2024semantic}
Xin Yang, Wending Yan, Yuan Yuan, Michael~Bi Mi, and Robby~T Tan.
\newblock Semantic segmentation in multiple adverse weather conditions with
  domain knowledge retention.
\newblock In \emph{Proceedings of the AAAI Conference on Artificial
  Intelligence}, pages 6558--6566, 2024.

\bibitem[Yang et~al.(2025)Yang, Wending, Bi~Mi, Yuan, and Tan]{yang2025end}
Xin Yang, Yan Wending, Michael Bi~Mi, Yuan Yuan, and Robby Tan.
\newblock End-to-end video semantic segmentation in adverse weather using
  fusion blocks and temporal-spatial teacher-student learning.
\newblock \emph{Advances in Neural Information Processing Systems},
  37:\penalty0 141000--141020, 2025.

\bibitem[Ye et~al.(2024)Ye, Oh, et~al.]{ye2024otseg}
Jong~Chul Ye, Yujin Oh, et~al.
\newblock Otseg: Multi-prompt sinkhorn attention for zero-shot semantic
  segmentation.
\newblock In \emph{The 18th European Conference on Computer Vision, ECCV 2024}.
  European Computer Vision Association (ECVA), 2024.

\bibitem[Yi et~al.(2024)Yi, Bi, Zheng, Zhan, Ji, Huang, Li, and
  Zheng]{yi2024learning}
Jingjun Yi, Qi Bi, Hao Zheng, Haolan Zhan, Wei Ji, Yawen Huang, Yuexiang Li,
  and Yefeng Zheng.
\newblock Learning spectral-decomposited tokens for domain generalized semantic
  segmentation.
\newblock In \emph{Proceedings of the 32nd ACM International Conference on
  Multimedia}, pages 8159--8168, 2024.

\bibitem[Yu et~al.(2020)Yu, Chen, Wang, Xian, Chen, Liu, Madhavan, and
  Darrell]{yu2020bdd100k}
Fisher Yu, Haofeng Chen, Xin Wang, Wenqi Xian, Yingying Chen, Fangchen Liu,
  Vashisht Madhavan, and Trevor Darrell.
\newblock Bdd100k: A diverse driving dataset for heterogeneous multitask
  learning.
\newblock In \emph{Proceedings of the IEEE/CVF conference on computer vision
  and pattern recognition}, pages 2636--2645, 2020.

\bibitem[Zhai et~al.(2023)Zhai, Mustafa, Kolesnikov, and
  Beyer]{zhai2023sigmoid}
Xiaohua Zhai, Basil Mustafa, Alexander Kolesnikov, and Lucas Beyer.
\newblock Sigmoid loss for language image pre-training.
\newblock In \emph{Proceedings of the IEEE/CVF International Conference on
  Computer Vision}, pages 11975--11986, 2023.

\bibitem[Zhao et~al.(2022)Zhao, Zhong, Zhao, Sebe, and Lee]{zhao2022style}
Yuyang Zhao, Zhun Zhong, Na Zhao, Nicu Sebe, and Gim~Hee Lee.
\newblock Style-hallucinated dual consistency learning for domain generalized
  semantic segmentation.
\newblock In \emph{European conference on computer vision}, pages 535--552.
  Springer, 2022.

\bibitem[Zhao et~al.(2023)Zhao, Zhong, Zhao, Sebe, and Lee]{zhao2023style}
Yuyang Zhao, Zhun Zhong, Na Zhao, Nicu Sebe, and Gim~Hee Lee.
\newblock {Style-hallucinated dual consistency learning: A unified framework
  for visual domain generalization}.
\newblock \emph{IJCV}, 2023.

\bibitem[Zhong et~al.(2022)Zhong, Zhao, Lee, and Sebe]{zhong2022adversarial}
Zhun Zhong, Yuyang Zhao, Gim~Hee Lee, and Nicu Sebe.
\newblock Adversarial style augmentation for domain generalized urban-scene
  segmentation.
\newblock \emph{Advances in neural information processing systems},
  35:\penalty0 338--350, 2022.

\bibitem[Zhou et~al.(2021)Zhou, Wei, Wang, Shen, Xie, Yuille, and
  Kong]{zhou2021ibot}
Jinghao Zhou, Chen Wei, Huiyu Wang, Wei Shen, Cihang Xie, Alan Yuille, and Tao
  Kong.
\newblock ibot: Image bert pre-training with online tokenizer.
\newblock \emph{arXiv preprint arXiv:2111.07832}, 2021.

\bibitem[Zhou et~al.(2023)Zhou, Lei, Zhang, Liu, and Liu]{zhou2023zegclip}
Ziqin Zhou, Yinjie Lei, Bowen Zhang, Lingqiao Liu, and Yifan Liu.
\newblock Zegclip: Towards adapting clip for zero-shot semantic segmentation.
\newblock In \emph{Proceedings of the IEEE/CVF Conference on Computer Vision
  and Pattern Recognition}, pages 11175--11185, 2023.

\bibitem[Zhu et~al.(2024)Zhu, Liao, Zhang, Wang, Liu, and Wang]{zhu2024vision}
Lianghui Zhu, Bencheng Liao, Qian Zhang, Xinlong Wang, Wenyu Liu, and Xinggang
  Wang.
\newblock Vision mamba: Efficient visual representation learning with
  bidirectional state space model.
\newblock \emph{arXiv preprint arXiv:2401.09417}, 2024.

\bibitem[Zhu et~al.(2020)Zhu, Su, Lu, Li, Wang, and Dai]{zhu2020deformable}
Xizhou Zhu, Weijie Su, Lewei Lu, Bin Li, Xiaogang Wang, and Jifeng Dai.
\newblock Deformable detr: Deformable transformers for end-to-end object
  detection.
\newblock \emph{arXiv preprint arXiv:2010.04159}, 2020.

\end{thebibliography}
}

\end{document}